# Graph Learning from Data under Structural and Laplacian Constraints

Hilmi E. Egilmez, Eduardo Pavez, and Antonio Ortega

***Abstract*—Graphs are fundamental mathematical structures used in various fields to represent data, signals and processes. In this paper, we propose a novel framework for learning/estimating graphs from data. The proposed framework includes (i) formulation of various graph learning problems, (ii) their probabilistic interpretations and (iii) associated algorithms. Specifically, graph learning problems are posed as estimation of graph Laplacian matrices from some observed data under given structural constraints (e.g., graph connectivity and sparsity level). From a probabilistic perspective, the problems of interest correspond to maximum a posteriori (MAP) parameter estimation of Gaussian-Markov random field (GMRF) models, whose precision (inverse covariance) is a graph Laplacian matrix. For the proposed graph learning problems, specialized algorithms are developed by incorporating the graph Laplacian and structural constraints. The experimental results demonstrate that the proposed algorithms outperform the current state-of-the-art methods in terms of accuracy and computational efficiency.**

## I. Introduction

GRAPHS are generic mathematical structures consisting of sets of vertices and edges, which are used for modeling pairwise relations (edges) between a number of objects (vertices). In practice, this representation is often extended to weighted graphs, for which a set of scalar values (weights) are assigned to edges and potentially to vertices. Thus, weighted graphs offer general and flexible representations for modeling affinity relations between the objects of interest.

Many practical problems can be represented using weighted graphs. For example, a broad class of combinatorial problems such as weighted matching, shortest-path and network-flow [2] are defined using weighted graphs. In signal/data-oriented problems, weighted graphs provide concise (sparse) representations for robust modeling of signals/data [3]. Such graph-based models are also useful for analyzing and visualizing the relations between their samples/features. Moreover, weighted graphs naturally emerge in networked data applications, such as learning, signal processing and analysis on computer, social, sensor, energy, transportation and biological networks [4], where the signals/data are inherently related to a graph associated with the underlying network. Similarly, image and video signals can be modeled using weighted graphs whose weights capture the correlation or similarity between neighboring pixel values (such as in nearest-neighbor models) [5]–[10]. In practice, many datasets typically consist of an unstructured list of samples, where the associated graph information (representing the structural relations between samples/features) is latent. For analysis, processing and algorithmic purposes, it is often useful to build graph-based models that provide a concise/simple explanation for datasets and also reduce the dimension of the problem by exploiting the available prior knowledge/assumptions about the desired graph (e.g., structural information including connectivity and sparsity level).

The focus of this paper is on learning graphs (i.e., graph-based models) from data, where the basic goal is to optimize a weighted graph with nonnegative edge weights characterizing the affinity relationship between the entries of a signal/data vector based on multiple observed vectors. For this purpose, we propose a general framework where graph learning is formulated as the estimation of different types of graph Laplacian matrices from data. Specifically, our formulations involve, for a given data statistic $\mathbf{S}$ (e.g., empirically obtained covariance/kernel matrices) and a symmetric regularization matrix $\mathbf{H}$, minimization of objective functions of the following form:

$$\underbrace{\operatorname{Tr}(\mathbf{\Theta S}) - \operatorname{logdet}(\mathbf{\Theta})}_{\mathcal{D}(\mathbf{\Theta},\mathbf{S})} + \underbrace{\|\mathbf{\Theta} \odot \mathbf{H}\|_1}_{\mathcal{R}(\mathbf{\Theta},\mathbf{H})}, \tag{1}$$

where $\mathbf{\Theta}$ denotes the $n \times n$ target matrix variable, $\mathcal{R}(\mathbf{\Theta},\mathbf{H})$ is the sparsity promoting weighted $\ell_1$-regularization term [11] multiplying $\mathbf{\Theta}$ and $\mathbf{H}$ element-wise, and $\mathcal{D}(\mathbf{\Theta},\mathbf{S})$ is the data-fidelity term, a log-determinant Bregman divergence [12], whose minimization corresponds to the maximum likelihood estimation of inverse covariance (precision) matrices for multivariate Gaussian distributions. Thus, minimizing (1) for arbitrary data can be interpreted as finding the parameters of a multivariate Gaussian model that best approximates the data [13] [14]. In addition to the objective in (1), our formulations incorporate problem-specific *Laplacian and structural constraints* depending on (i) the desired type of graph Laplacian and (ii) the available structural information about the graph structure. Particularly, we consider three types of graph Laplacian matrices, namely, *generalized graph Laplacians (GGLs)*, *diagonally dominant generalized graph Laplacians (DDGLs)*, and *combinatorial graph Laplacians (CGLs)*, and develop novel techniques to estimate them from data (i.e., data statistic $\mathbf{S}$). As illustrated in Fig. 1 and further discussed in Section IV, the proposed graph Laplacian estimation techniques can also be viewed as methods to learn different classes of Gaussian-Markov random fields (GMRFs) [15] [16], whose precision

Authors are with the Department of Electrical Engineering, University of Southern California, Los Angeles, CA, 90089 USA. Contact author e-mail: hegilmez@usc.edu. This work is supported in part by NSF under grants CCF-1410009 and CCF-1527874. An earlier version of this paper is presented at 50th Asilomar Conference on Signals, Systems and Computers, 2016 [1].

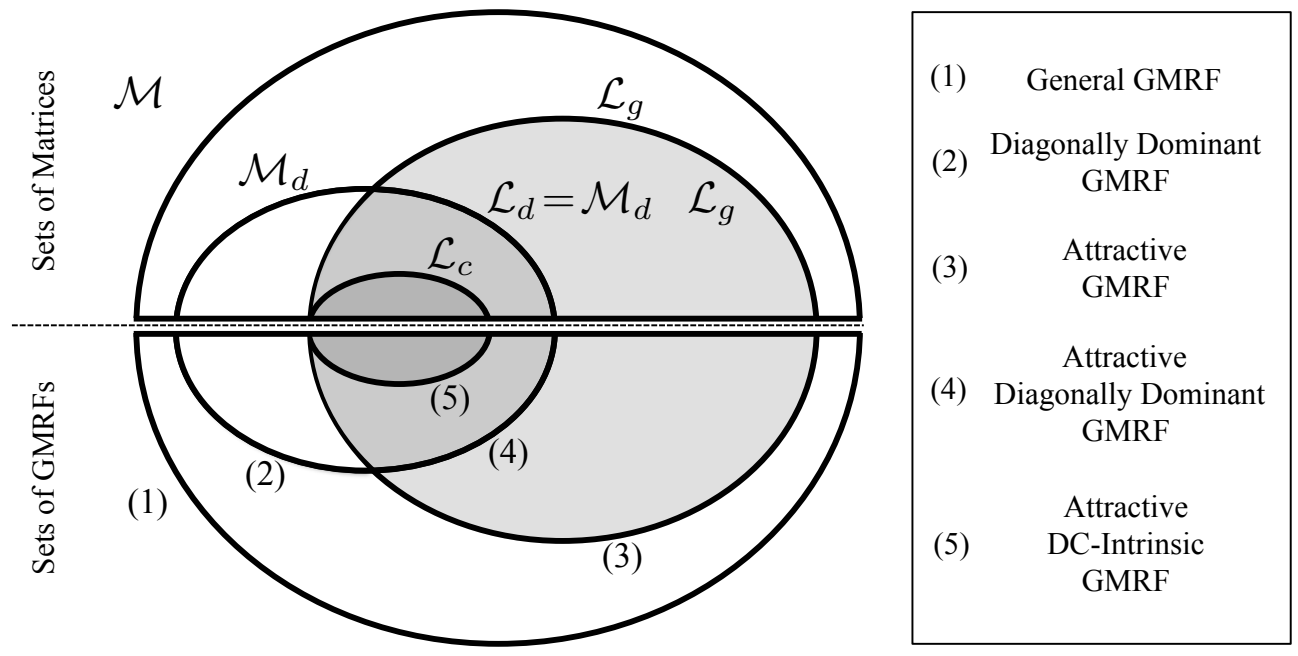

Fig. 1. The set of positive semidefinite matrices ($\mathcal{M}_{\text{psd}}$) containing the sets of diagonally dominant positive semidefinite matrices ($\mathcal{M}_d$), generalized ($\mathcal{L}_g$), diagonally dominant ($\mathcal{L}_d$) and combinatorial Laplacian ($\mathcal{L}_c$) matrices. The corresponding classes of GMRFs are enumerated as (1)–(5), respectively. In this work, we focus on estimating/learning the sets colored in gray.

matrices are graph Laplacians. Moreover, in our formulations, structural (connectivity) constraints are introduced to exploit available prior information about the target graph. When graph connectivity is unknown, graph learning involves estimating both graph structure and graph weights, with the regularization term controlling the level of sparsity. Otherwise, if graph connectivity is given (e.g., based on application-specific assumptions or prior knowledge), graph learning reduces to the estimation of graph weights only.

Graph Laplacian matrices have multiple applications in various fields. In spectral graph theory [17], basic properties of graphs are investigated by analyzing characteristic polynomials, eigenvalues and eigenvectors of the associated graph Laplacian matrices. In machine learning, graph Laplacians are extensively used as kernels, especially in spectral clustering [18] and graph regularization [19] tasks. Moreover, in graph signal processing [4], basic signal processing operations such as filtering [10] [20], sampling [21], transformation [9] [22] are extended to signals defined on graphs associated with Laplacian matrices. Although the majority of studies and applications primarily focus on CGLs (and their normalized versions) [17] [23], which represent graphs with zero vertex weights (i.e., graphs with no self-loops), there are recent studies where GGLs [24] (i.e., graphs with nonzero vertex weights) are shown to be useful. Particularly, GGLs are proposed for modeling image and video signals in [8] [9], and their potential machine learning applications are discussed in [25]. In [26], a Kron reduction procedure is developed based on GGLs for simplified modeling of electrical networks. Furthermore, DDGLs are utilized in [27]–[29] to develop efficient algorithms for graph partitioning [27], graph sparsification [28] and solving linear systems [29]. Our work complements these methods and applications by proposing efficient algorithms for estimation of these three types of graph Laplacians from data.

In the literature, several approaches have been proposed for estimating graph-based models. Dempster [30] originally proposed the idea of introducing zero entries in inverse covariance matrices for simplified covariance estimation. Later, a neighborhood selection approach was proposed for graphical model estimation [31] by using the Lasso algorithm [32]. Friedman *et al.* [33] formulated a regularization framework for sparse inverse covariance estimation and developed the Graphical Lasso algorithm to solve the regularized problem. Some algorithmic extensions of the Graphical Lasso are discussed in [13], [34], and a few computationally efficient variations are presented in [35]–[37]. However, inverse covariance estimation methods, such as the Graphical Lasso, search for solutions in the set of the positive semidefinite matrices ($\mathcal{M}_{\text{psd}}$ in Fig. 1), which lead to a different notion of graphs by allowing both negative and positive edge weights, while we focus on learning graphs with nonnegative edge weights, associated with graph Laplacian matrices ($\mathcal{L}_g$, $\mathcal{L}_d$ or $\mathcal{L}_c$ in Fig. 1). Although graph Laplacian matrices represent a more restricted set of models (attractive GMRFs) compared to positive semidefinite matrices (modeling general GMRFs), attractive GMRFs cover an important class of random vectors whose entries can be optimally predicted by nonnegative linear combinations of the other entries. For this class of signals/data, our proposed algorithms incorporating Laplacian constraints provide more accurate graph estimation than sparse inverse covariance methods (e.g., Graphical Lasso). Even when such model assumptions do not strictly hold, the proposed algorithms can be employed to find the best (closest) graph Laplacian fit with respect to the Bregman divergence in (1) for applications where graph Laplacians are useful (e.g., in [4], [9], [17]–[29]).

This paper proposes a general optimization framework for estimating graph Laplacian matrices by (i) introducing new problem formulations with Laplacian and structural (i.e., connectivity) constraints and (ii) developing novel algorithms specialized for the proposed problems, which provide more accurate and efficient graph estimation compared to the current state-of-the-art approaches.

Several recent publications address learning of different types of graph Laplacians from data. Closest to our work, Slawski and Hein address the problem of estimating symmetric M-matrices [38], or equivalently GGLs, and propose an efficient primal algorithm [39], while our recent work [40] proposes an alternative dual algorithm for GGL estimation. Our work in this paper addresses the same GGL estimation problem as [39] [40], based on a primal approach analogous to that of [39], but unlike both [39] and [40], we incorporate connectivity constraints in addition to sparsity promoting regularization. For estimation of CGLs, Lake and Tenenbaum [41] also consider minimization of the objective function in (1), which is unbounded for CGLs (since they are singular matrices). To avoid working with singular matrices, they propose to optimize a different target matrix obtained by adding a positive constant value to diagonal entries of a combinatorial Laplacian, but no efficient algorithm is developed. Dong *et al.* [42] and Kalofolias [43] propose minimization of two objective functions different from (1) in order to overcome issues related to the singularity of CGLs. Instead, by restricting our learning problem to connected graphs (which have exactly one eigenvector with eigenvalue 0), we can directly use a modified version of (1) as the objective function and develop an efficient algorithm that guarantees convergence to the optimal solution, with significant improvements in experimental results over prior work.

There are also a few recent studies that focus on inferring graph topology (i.e., connectivity) information from signals assumed to be diffused on a graph. Particularly, Segarra *et al.* [44] and Pasdeloup *et al.* [45] focus on learning graph shift/diffusion operators (such as adjacency and Laplacian matrices) from a set of diffused graph signals, and Sardellitti *et al.* [46] propose an approach to estimate a graph Laplacian from bandlimited graph signals. None of these works [44]–[46] considers the minimization of (1). In fact, techniques proposed in these papers directly use the eigenvectors of the empirical covariance matrix and only optimize the choice of eigenvalues of the Laplacian or adjacency matrices under specific criteria, for the given eigenvectors. In contrast, our methods implicitly optimize both eigenvectors and eigenvalues by minimizing (1). Addressing specifically the problem of learning diffusion-based models as in [44]–[46] is out of the scope of this paper and will be considered in our future work.

**Summary of contributions.** We address estimation of three different types of graph Laplacian with structural constraints. To the best of our knowledge, the estimation of DDGLs is considered for the first time in this paper. For CGL estimation, we propose a novel formulation for the objective function in (1), whose direct minimization is not possible due to the singularity of CGLs. Our formulation allows us to improve the accuracy of CGL estimation significantly compared to the approaches in [41]–[43]. For GGL estimation, the prior formulations in [39][40] are extended in order to accommodate structural constraints. To solve the proposed problems, we develop efficient block-coordinate descent (BCD) algorithms [47] exploiting the structural constraints within the problems, which can significantly improve the accuracy and reduce the computational complexity depending on the degree of sparsity introduced by the constraints. Moreover, we theoretically show that the proposed algorithms guarantee convergence to the optimal solution. Previously, numerous BCD-type algorithms are proposed for sparse inverse covariance estimation [13][33][34] which iteratively solve an $\ell_1$-regularized quadratic program. However, our algorithms are specifically developed for graph Laplacian estimation problems, where we solve a nonnegative quadratic program for block-coordinate updates. Finally, we present probabilistic interpretations of our proposed problems by showing that their solutions lead to optimal parameter estimation for special classes of GMRFs, as depicted in Fig. 1. While recent work has noted the relation between graph Laplacians and GMRFs [7] [48], the present paper provides a more comprehensive classification of GMRFs and proposes specific methods for estimation of their parameters.

The rest of the paper is organized as follows. In Section II, we present the notations and basic concepts used throughout the paper. Section III formulates our proposed problems and summarizes some of the related formulations in the literature. Section IV discusses the probabilistic interpretation of our proposed problems. In Section V, we derive necessary and sufficient optimality conditions and develop novel algorithms for the proposed graph learning problems. Experimental results are presented in Section VI, and some concluding remarks are discussed in Section VII. Our proofs for the propositions stated throughout the paper are included in the appendix.

TABLE I
LIST OF SYMBOLS AND THEIR MEANING

| Symbols | Meaning |
|---|---|
| $\mathcal{G} \mid \mathbf{L}$ | weighted graph \| graph Laplacian matrix |
| $\mathcal{V} \mid \mathcal{E} \mid \mathcal{S}^c$ | vertex set \| edge set \| complement of set $\mathcal{S}$ |
| $\mathcal{P}_u$ | set of unordered pairs of vertices |
| $n \mid k$ | number of vertices \| number of data samples |
| $\mathbf{O} \mid \mathbf{I}$ | matrix of zeros \| identity matrix |
| $\mathbf{0} \mid \mathbf{1}$ | column vector of zeros \| column vector of ones |
| $\mathbf{W} \mid \mathbf{A}$ | adjacency matrix \| connectivity matrix |
| $\mathbf{D} \mid \mathbf{V}$ | degree matrix \| self-loop matrix |
| $\mathbf{H} \mid \alpha$ | regularization matrix \| regularization parameter |
| $\mathbf{\Theta}^{-1} \mid \mathbf{\Theta}^{\dagger}$ | inverse of $\mathbf{\Theta}$ \| pseudo-inverse of $\mathbf{\Theta}$ |
| $\mathbf{\Theta}^{\mathsf{T}} \mid \boldsymbol{\theta}^{\mathsf{T}}$ | transpose of $\mathbf{\Theta}$ \| transpose of $\boldsymbol{\theta}$ |
| $\det(\mathbf{\Theta}) \mid \lvert\mathbf{\Theta}\rvert$ | determinant of $\mathbf{\Theta}$ \| pseudo-determinant of $\mathbf{\Theta}$ |
| $(\mathbf{\Theta})_{ij}$ | entry of $\mathbf{\Theta}$ at $i$-th row and $j$-th column |
| $(\mathbf{\Theta})_{\mathcal{S}\mathcal{S}}$ | submatrix of $\mathbf{\Theta}$ formed by selecting indexes in $\mathcal{S}$ |
| $(\boldsymbol{\theta})_i$ | $i$-th entry of $\boldsymbol{\theta}$ |
| $(\boldsymbol{\theta})_{\mathcal{S}}$ | subvector of $\boldsymbol{\theta}$ formed by selecting indexes in $\mathcal{S}$ |
| $\geq$ $(\leq)$ | elementwise greater (less) than or equal to operator |
| $\mathbf{\Theta} \succeq 0$ | $\mathbf{\Theta}$ is a positive semidefinite matrix |
| $\mathrm{Tr} \mid \mathrm{logdet}(\mathbf{\Theta})$ | trace operator \| natural logarithm of $\det(\mathbf{\Theta})$ |
| $\mathrm{diag}(\boldsymbol{\theta})$ | diagonal matrix formed by elements of $\boldsymbol{\theta}$ |
| $\mathrm{ddiag}(\mathbf{\Theta})$ | diagonal matrix formed by diagonal elements of $\mathbf{\Theta}$ |
| $\mathbf{x} \sim \mathsf{N}(\mathbf{0}, \mathbf{\Sigma})$ | zero-mean multivariate Gaussian with covariance $\mathbf{\Sigma}$ |
| $\lVert\boldsymbol{\theta}\rVert_1$, $\lVert\mathbf{\Theta}\rVert_1$ | sum of absolute values of all elements ($\ell_1$-norm) |
| $\lVert\mathbf{\Theta}\rVert_{1,\mathrm{off}}$ | sum of absolute values of all off-diagonal elements |
| $\lVert\boldsymbol{\theta}\rVert_2^2$, $\lVert\mathbf{\Theta}\rVert_F^2$ | sum of squared values of all elements |
| $\lVert\mathbf{\Theta}\rVert_{F,\mathrm{off}}^2$ | sum of squared values of all off-diagonal elements |

## II. NOTATION AND PRELIMINARIES

### A. Notation

Throughout the paper, lowercase normal (e.g., $a$ and $\theta$), lowercase bold (e.g., $\mathbf{a}$ and $\boldsymbol{\theta}$) and uppercase bold (e.g., $\mathbf{A}$ and $\mathbf{\Theta}$) letters denote scalars, vectors and matrices, respectively. Unless otherwise stated, calligraphic capital letters (e.g., $\mathcal{E}$ and $\mathcal{S}$) represent sets. $O(\cdot)$ and $\Omega(\cdot)$ are the standard big-O and big-Omega notations used in complexity theory [2]. The rest of the notation is presented in Table I.

### B. Basic Definitions

We present basic definitions related to graphs. In this paper, we restrict our attention to undirected, weighted graphs with nonnegative edge weights.

**Definition 1** (Weighted Graph)**.** The graph $\mathcal{G} = (\mathcal{V}, \mathcal{E}, f_w, f_v)$ is a weighted graph with $n$ vertices in the set $\mathcal{V} = \{v_1, ..., v_n\}$. The edge set $\mathcal{E} = \{\, e \mid f_w(e) \neq 0, \forall\, e \in \mathcal{P}_u \}$ is a subset of $\mathcal{P}_u$, the set of all unordered pairs of distinct vertices, where $f_w((v_i, v_j)) \geq 0$ for $i \neq j$ is a real-valued edge weight function, and $f_v(v_i)$ for $i = 1, ..., n$ is a real-valued vertex (self-loop) weight function.

**Definition 2** (Simple Weighted Graph)**.** A simple weighted graph is a weighted graph with no self-loops (i.e., $f_v(v_i) = 0$ for $i = 1, ..., n$).

Weighted graphs can be represented by adjacency, degree and self-loop matrices, which are used to define graph Laplacian matrices. Moreover, we use connectivity matrices to incorporate structural constraints in our formulations. In the following, we present formal definitions for these matrices.

**Definition 3** (Algebraic representations of graphs)**.** For a given weighted graph $\mathcal{G}=(\mathcal{V},\mathcal{E},f_w,f_v)$ with $n$ vertices, $v_1,...,v_n$:
The **adjacency matrix** of $\mathcal{G}$ is an $n\times n$ symmetric matrix, $\mathbf{W}$, such that $(\mathbf{W})_{ij}=(\mathbf{W})_{ji}=f_w((v_i,v_j))$ for $(v_i,v_j)\in\mathcal{P}_u$.
The **degree matrix** of $\mathcal{G}$ is an $n\times n$ diagonal matrix, $\mathbf{D}$, with entries $(\mathbf{D})_{ii}=\sum_{j=1}^{n}(\mathbf{W})_{ij}$ and $(\mathbf{D})_{ij}=0$ for $i\neq j$.
The **self-loop matrix** of $\mathcal{G}$ is an $n\times n$ diagonal matrix, $\mathbf{V}$, with entries $(\mathbf{V})_{ii}=f_v(v_i)$ for $i=1,...,n$ and $(\mathbf{V})_{ij}=0$ for $i\neq j$. If $\mathcal{G}$ is a simple weighted graph, then $\mathbf{V}=\mathbf{O}$.
The **connectivity matrix** of $\mathcal{G}$ is an $n\times n$ matrix, $\mathbf{A}$, such that $(\mathbf{A})_{ij}=1$ if $(\mathbf{W})_{ij}\neq 0$, and $(\mathbf{A})_{ij}=0$ if $(\mathbf{W})_{ij}=0$ for $i,j=1,...,n$, where $\mathbf{W}$ is the adjacency matrix of $\mathcal{G}$.
The **generalized graph Laplacian** of a weighted graph $\mathcal{G}$ is defined as $\mathbf{L}=\mathbf{D}-\mathbf{W}+\mathbf{V}$.
The **combinatorial graph Laplacian** of a simple weighted graph $\mathcal{G}$ is defined as $\mathbf{L}=\mathbf{D}-\mathbf{W}$.

**Definition 4** (Diagonally Dominant Matrix)**.** An $n\times n$ matrix $\boldsymbol{\Theta}$ is diagonally dominant if $|(\boldsymbol{\Theta})_{ii}|\geq\sum_{j\neq i}|(\boldsymbol{\Theta})_{ij}|\ \forall i$, and it is strictly diagonally dominant if $|(\boldsymbol{\Theta})_{ii}|>\sum_{j\neq i}|(\boldsymbol{\Theta})_{ij}|\ \forall i$.

Note that any weighted graph with positive edge weights (see Definitions 1 and 2) can be represented by a generalized graph Laplacian, while simple weighted graphs lead to combinatorial graph Laplacians, since they have no self-loops (i.e., $\mathbf{V}=\mathbf{O}$). Moreover, if a weighted graph has nonnegative vertex weights (i.e., $\mathbf{V}\geq\mathbf{O}$), its generalized Laplacian matrix is also diagonally dominant.

Based on the definitions above, the sets of graph Laplacian matrices considered in this paper can be written as

- $\mathcal{L}_g=\{\mathbf{L}\,|\,\mathbf{L}\succeq 0,\ (\mathbf{L})_{ij}\leq 0 \text{ for } i\neq j\}$,
- $\mathcal{L}_d=\{\mathbf{L}\,|\,\mathbf{L}\succeq 0,\ (\mathbf{L})_{ij}\leq 0 \text{ for } i\neq j,\ \mathbf{L}\mathbf{1}\geq\mathbf{0}\}$,
- $\mathcal{L}_c=\{\mathbf{L}\,|\,\mathbf{L}\succeq 0,\ (\mathbf{L})_{ij}\leq 0 \text{ for } i\neq j,\ \mathbf{L}\mathbf{1}=\mathbf{0}\}$,

which are illustrated in Fig.1.

## III. Problem Formulations for Graph Learning

### A. Proposed Formulations: Graph Laplacian Estimation

Assume that we are given a $k\times n$ data matrix $\mathbf{X}$ (where each column corresponds to a vertex), and the goal is to estimate a graph Laplacian matrix from a data statistic, $\mathbf{S}$. Depending on the application and underlying statistical assumptions, $\mathbf{S}$ may stand for the sample covariance of $\mathbf{X}$ or a kernel matrix $\mathbf{S}=\mathcal{K}(\mathbf{X},\mathbf{X})$ derived from data, where $\mathcal{K}$ is a positive definite kernel function (e.g., polynomial and RBF kernels). For the purpose of graph learning, we formulate three different optimization problems for a given $\mathbf{S}$, a connectivity matrix $\mathbf{A}$ and a regularization matrix $\mathbf{H}$. In our problems, we minimize the function in (1) under the following set of constraints defined based on a given set of graph Laplacians $\mathcal{L}$ and a connectivity matrix $\mathbf{A}$,

$$\mathcal{L}(\mathbf{A})=\left\{\boldsymbol{\Theta}\in\mathcal{L}\ \middle|\ \begin{array}{ll}(\boldsymbol{\Theta})_{ij}\leq 0 & \text{if } (\mathbf{A})_{ij}=1\\ (\boldsymbol{\Theta})_{ij}=0 & \text{if } (\mathbf{A})_{ij}=0\end{array}\ \text{for } i\neq j\right\}. \quad (2)$$

Based on the nonpositivity constraints on $\boldsymbol{\Theta}$ (i.e., nonnegativity of edge weights), a regularization matrix $\mathbf{H}$ can be selected such that $\mathcal{R}(\boldsymbol{\Theta},\mathbf{H})$ term in (1) is compactly written as

$$\|\boldsymbol{\Theta}\odot\mathbf{H}\|_1=\mathrm{Tr}\,(\boldsymbol{\Theta}\mathbf{H}). \quad (3)$$

For example, the following standard $\ell_1$-regularization terms with parameter $\alpha$ can be written in the above form as,

$$\alpha\|\boldsymbol{\Theta}\|_1=\mathrm{Tr}\,(\boldsymbol{\Theta}\mathbf{H})\ \text{ where } \mathbf{H}=\alpha(2\mathbf{I}-\mathbf{1}\mathbf{1}^{\mathsf{T}}), \quad (4)$$

$$\alpha\|\boldsymbol{\Theta}\|_{1,\text{off}}=\mathrm{Tr}\,(\boldsymbol{\Theta}\mathbf{H})\ \text{ where } \mathbf{H}=\alpha(\mathbf{I}-\mathbf{1}\mathbf{1}^{\mathsf{T}}). \quad (5)$$

Since the trace operator is linear, we can rewrite the objective function in (1) as

$$\mathrm{Tr}\,(\boldsymbol{\Theta}\mathbf{K})-\mathrm{logdet}(\boldsymbol{\Theta})\ \text{ where } \mathbf{K}=\mathbf{S}+\mathbf{H}, \quad (6)$$

which is the form used in our optimization problems. Note that the nonnegativity of edge weights allows us to transform the nonsmooth function in (1) into the smooth function in (6) by rewriting the regularization term as in (3).

**Problem 1** (GGL Problem)**.** The optimization problem formulated for estimating generalized graph Laplacian (GGL) matrices is

$$\begin{array}{ll}\underset{\boldsymbol{\Theta}}{\text{minimize}} & \mathrm{Tr}\,(\boldsymbol{\Theta}\mathbf{K})-\mathrm{logdet}(\boldsymbol{\Theta})\\ \text{subject to} & \boldsymbol{\Theta}\in\mathcal{L}_g(\mathbf{A})\end{array} \quad (7)$$

where $\mathbf{K}=\mathbf{S}+\mathbf{H}$ as in (6), and the set of constraints $\mathcal{L}_g(\mathbf{A})$ leads to $\boldsymbol{\Theta}$ being a GGL matrix.

**Problem 2** (DDGL Problem)**.** The diagonally dominant generalized graph Laplacian (DDGL) estimation problem is formulated as

$$\begin{array}{ll}\underset{\boldsymbol{\Theta}}{\text{minimize}} & \mathrm{Tr}\,(\boldsymbol{\Theta}\mathbf{K})-\mathrm{logdet}(\boldsymbol{\Theta})\\ \text{subject to} & \boldsymbol{\Theta}\in\mathcal{L}_d(\mathbf{A})\end{array} \quad (8)$$

where the additional $\boldsymbol{\Theta}\mathbf{1}\geq\mathbf{0}$ constraint in $\mathcal{L}_d(\mathbf{A})$ ensures that all vertex weights are nonnegative, and therefore the optimal solution is a diagonally dominant matrix.

**Problem 3** (CGL Problem)**.** The combinatorial graph Laplacian (CGL) estimation problem is formulated as

$$\begin{array}{ll}\underset{\boldsymbol{\Theta}}{\text{minimize}} & \mathrm{Tr}\,(\boldsymbol{\Theta}\mathbf{K})-\log|\boldsymbol{\Theta}|\\ \text{subject to} & \boldsymbol{\Theta}\in\mathcal{L}_c(\mathbf{A})\end{array} \quad (9)$$

where the objective function involves the pseudo-determinant term ($|\boldsymbol{\Theta}|$), since the target matrix $\boldsymbol{\Theta}$ is singular. However, the problem is hard to solve because of the $|\boldsymbol{\Theta}|$ term. To cope with this, we propose to reformulate (9) as the following problem[1],

$$\begin{array}{ll}\underset{\boldsymbol{\Theta}}{\text{minimize}} & \mathrm{Tr}\,(\boldsymbol{\Theta}(\mathbf{K}+\mathbf{J}))-\mathrm{logdet}(\boldsymbol{\Theta}+\mathbf{J})\\ \text{subject to} & \boldsymbol{\Theta}\in\mathcal{L}_c(\mathbf{A})\end{array} \quad (10)$$

where the $\boldsymbol{\Theta}\mathbf{1}=\mathbf{0}$ constraint in $\mathcal{L}_c(\mathbf{A})$ guarantees that the solution is a CGL matrix, and $\mathbf{J}=\mathbf{u}_1\mathbf{u}_1^{\mathsf{T}}=(1/n)\,\mathbf{1}\mathbf{1}^{\mathsf{T}}$ such that $\mathbf{u}_1$ is the eigenvector corresponding to the zero eigenvalue of CGL matrices.

**Proposition 1.** *The optimization problems stated in (9) and (10) are equivalent.*

**Proposition 2.** *Problems 1, 2 and 3 are convex optimization problems.*

[1]An alternative second-order approach is proposed to solve (10) in [49], which is published after the initial submission of the present paper [50]. Yet, the equivalence of (9) and (10) is not discussed in [49].

In Problems 1, 2 and 3, prior knowledge/assumptions about the graph structure are built into the choice of $\mathbf{A}$, determining the structural constraints. In practice, if the graph connectivity is unknown, then $\mathbf{A}$ can be set to represent a fully connected graph, $\mathbf{A}=\mathbf{A}_{\text{full}}=\mathbf{1}\mathbf{1}^{\mathsf{T}}-\mathbf{I}$, and the regularization matrix $\mathbf{H}$ (or the parameter $\alpha$ for the $\ell_1$-regularizations in (4) and (5)) can be tuned until the desired level of sparsity is achieved.

### B. Related Prior Formulations

**Sparse Inverse Covariance Estimation** [33]**.** The goal is to estimate a sparse inverse covariance matrix from $\mathbf{S}$ by solving:

$$\underset{\mathbf{\Theta}\succeq 0}{\text{minimize}} \;\; \mathrm{Tr}\left(\mathbf{\Theta}\mathbf{S}\right) - \mathrm{logdet}\left(\mathbf{\Theta}\right) + \alpha\|\mathbf{\Theta}\|_1. \tag{11}$$

In this paper, we are interested in minimization of the same objective function under Laplacian and structural constraints.

**Shifted CGL Estimation** [41]**.** The goal is to estimate a shifted CGL matrix, which is defined by adding a scalar value to diagonal entries of a combinatorial Laplacian matrix:

$$\begin{aligned} \underset{\mathbf{\Theta}\succeq 0,\,\nu\geq 0}{\text{minimize}} \quad & \mathrm{Tr}\left(\mathbf{\Theta}\mathbf{S}\right) - \mathrm{logdet}\left(\mathbf{\Theta}\right) + \alpha\|\mathbf{\Theta}\|_1 \\ \text{subject to} \quad & \mathbf{\Theta} = \widetilde{\mathbf{\Theta}} + \nu\mathbf{I} \\ & \widetilde{\mathbf{\Theta}}\mathbf{1} = \mathbf{0},\; (\widetilde{\mathbf{\Theta}})_{ij} \leq 0 \quad i \neq j \end{aligned} \tag{12}$$

where $\nu$ denotes the positive scalar (i.e., shift) variable added to diagonal elements of $\widetilde{\mathbf{\Theta}}$, which is constrained to be a CGL matrix, so that $\mathbf{\Theta}$ is the target variable. By solving this problem, a CGL matrix $\widehat{\mathbf{\Theta}}$ is estimated by subtracting the shift variable as $\widehat{\mathbf{\Theta}}=(\mathbf{\Theta}-\nu\mathbf{I})$. However, this generally leads to a different solution than our method (see Appendix for the proof of the following proposition).

**Proposition 3.** *The objective functions of Problem 3 and the shifted CGL problem in (12) are different.*

**Graph Learning from Smooth Signals** [42] [43]**.** The goal is to estimate a CGL from $n$-dimensional signals that are assumed to be smooth with respect to the corresponding graph:

$$\begin{aligned} \underset{\mathbf{\Theta}\succeq 0}{\text{minimize}} \;\; & \mathrm{Tr}\left(\mathbf{\Theta}\mathbf{S}\right) + \alpha_1\|\mathbf{\Theta}\|_F^2 \\ \text{subject to} \;\; & \mathbf{\Theta}\mathbf{1} = \mathbf{0},\; \mathrm{Tr}\left(\mathbf{\Theta}\right) = n,\; (\mathbf{\Theta})_{ij} \leq 0 \quad i \neq j \end{aligned} \tag{13}$$

where the sum of degrees is constrained as $\mathrm{Tr}\left(\mathbf{\Theta}\right) = n$. This is a limitation that is later relaxed in [43] by introducing the following problem with regularization parameters

$$\begin{aligned} \underset{\mathbf{\Theta}\succeq 0}{\text{minimize}} \;\; & \mathrm{Tr}\left(\mathbf{\Theta}\mathbf{S}\right) + \alpha_1\|\mathbf{\Theta}\|_{F,\text{off}}^2 - \alpha_2\sum_{i=1}^{n}\log\left((\mathbf{\Theta})_{ii}\right) \\ \text{subject to} \;\; & \mathbf{\Theta}\mathbf{1} = \mathbf{0}, \quad (\mathbf{\Theta})_{ij} \leq 0 \quad i \neq j \end{aligned} \tag{14}$$

where the constraints in (13) and (14) lead to a CGL solution. The following proposition relates the objective function in (14) with $\alpha_1 = 0$ to the objective in our proposed CGL estimation problem (see Appendix for a proof).

**Proposition 4.** *The objective function in Problem 3 with $\alpha = 0$ is lower-bounded by the objective function in (14) for $\alpha_1 = 0$ and $\alpha_2 = 1$.*

**Graph Topology Inference.** Various approaches for graph topology (connectivity) inference from data (under diffusion-based model assumptions) have been proposed in [44]–[46]. As the most related to our work, Segarra *et al.* [44] introduce a sparse recovery problem to infer the graph topology information from the eigenbasis, $\mathbf{U}$, associated with a graph shift/diffusion operator. Specifically for CGL estimation, the following problem is formulated:

$$\begin{aligned} \underset{\mathbf{\Theta}\succeq 0,\mathbf{\Lambda}}{\text{minimize}} \quad & \|\mathbf{\Theta}\|_1 \\ \text{subject to} \quad & \mathbf{\Theta} = \mathbf{U}\mathbf{\Lambda}\mathbf{U}^{\mathsf{T}} \\ & \mathbf{\Theta}\mathbf{1} = \mathbf{0},\; (\mathbf{\Theta})_{ij} \leq 0 \quad i \neq j \end{aligned} \tag{15}$$

where the eigenbasis $\mathbf{U}$ is the input to the problem, so that the goal is to find the set of eigenvalues (i.e., the diagonal matrix $\mathbf{\Lambda}$) minimizing $\|\mathbf{\Theta}\|_1$. Note that the problems in [44]–[46] require that $\mathbf{U}$ be given (or calculated beforehand), while our goal is to directly estimate a graph Laplacian so that both $\mathbf{U}$ and $\mathbf{\Lambda}$ are jointly optimized.

## IV. Probabilistic Interpretation of Proposed Graph Learning Problems

The proposed graph learning problems can be viewed from a probabilistic perspective by assuming that the data has been sampled from a zero-mean $n$-variate Gaussian distribution[2] $\mathbf{x} \sim \mathsf{N}(\mathbf{0}, \mathbf{\Sigma}=\mathbf{\Omega}^{\dagger})$, parametrized with a positive semidefinite precision matrix $\mathbf{\Omega}$, defining a Gaussian Markov random field (GMRF)

$$\mathsf{p}(\mathbf{x}|\mathbf{\Omega}) = \frac{1}{(2\pi)^{n/2}|\mathbf{\Omega}^{\dagger}|^{1/2}}\exp\left(-\frac{1}{2}\mathbf{x}^{\mathsf{T}}\mathbf{\Omega}\mathbf{x}\right), \tag{16}$$

with covariance matrix $\mathbf{\Sigma}=\mathbf{\Omega}^{\dagger}$. Based on its precision matrix ($\mathbf{\Omega}$), a GMRF is classified as [15] [16]:

- a *general GMRF* if its precision $\mathbf{\Omega}$ is positive semidefinite,
- an *attractive GMRF* if its precision $\mathbf{\Omega}$ has nonpositive off-diagonal entries,
- a *diagonally dominant GMRF* if its precision $\mathbf{\Omega}$ is diagonally dominant,
- an *intrinsic GMRF* if its precision $\mathbf{\Omega}$ is positive semidefinite and singular.

The entries of the precision matrix $\mathbf{\Omega}$ can be interpreted in terms of the following conditional dependence relations among the variables in $\mathbf{x}$,

$$\mathsf{E}\left[x_i \,|\, (\mathbf{x})_{\mathcal{S}\setminus\{i\}}\right] = -\frac{1}{(\mathbf{\Omega})_{ii}}\sum_{j\in\mathcal{S}\setminus\{i\}}(\mathbf{\Omega})_{ij}x_j \tag{17}$$

$$\mathsf{Prec}\left[x_i \,|\, (\mathbf{x})_{\mathcal{S}\setminus\{i\}}\right] = (\mathbf{\Omega})_{ii} \tag{18}$$

$$\mathsf{Corr}\left[x_i x_j \,|\, (\mathbf{x})_{\mathcal{S}\setminus\{i,j\}}\right] = -\frac{(\mathbf{\Omega})_{ij}}{\sqrt{(\mathbf{\Omega})_{ii}(\mathbf{\Omega})_{jj}}} \quad i \neq j, \tag{19}$$

where $\mathcal{S} = \{1, ..., n\}$ is the index set for $\mathbf{x} = [x_1, ..., x_n]^{\mathsf{T}}$. The conditional expectation in (17) represents the minimum mean square error (MMSE) prediction of $x_i$ using all the other random variables in $\mathbf{x}$. The *precision* of $x_i$ is defined as in (18), and the relation in (19) corresponds to the *partial correlation* between $x_i$ and $x_j$ (i.e., correlation between random variables $x_i$ and $x_j$ given all the other variables in $\mathbf{x}$). For example, if $x_i$ and $x_j$ are conditionally independent ($(\mathbf{\Omega})_{ij}=0$), there is no edge between corresponding vertices $v_i$ and $v_j$. For GMRFs, whose precision matrices are graph Laplacian matrices (i.e.,

[2]The zero-mean assumption is made to simplify the notation. Our analysis can be trivially extended to a multivariate Gaussian with nonzero mean.

$\mathbf{\Omega}=\mathbf{L}$), we can show that there is a one-to-one correspondence (bijection) between different classes of attractive GMRFs and types of graph Laplacian matrices by their definitions, as illustrated in Fig. 1:

- $\mathbf{L}$ is a GGL matrix ($\mathbf{L}\in\mathcal{L}_g$) if and only if $\mathsf{p}(\mathbf{x}|\mathbf{L})$ is an attractive GMRF,
- $\mathbf{L}$ is a DDGL matrix ($\mathbf{L}\in\mathcal{L}_d$) if and only if $\mathsf{p}(\mathbf{x}|\mathbf{L})$ is an attractive, diagonally dominant GMRF,
- $\mathbf{L}$ is a CGL matrix ($\mathbf{L}\in\mathcal{L}_c$) if and only if $\mathsf{p}(\mathbf{x}|\mathbf{L})$ is an attractive, DC-intrinsic GMRF.

Note that, in our characterization, the GMRFs corresponding to CGL matrices are classified as DC-intrinsic GMRFs, which are specific cases of intrinsic GMRFs [15] with no probability density along the direction of the eigenvector $\mathbf{u}_1 = 1/\sqrt{n}\,\mathbf{1}$ associated with the zero eigenvalue ($\lambda_1(\mathbf{L})=0$). On the other hand, if $\mathbf{L}$ is a nonsingular GGL matrix, then $\mathbf{x}$ has a proper (non-degenerate) distribution.

Moreover, for $\mathbf{\Omega}=\mathbf{L}$, the quadratic term in the exponent in (16) can be interpreted in terms of edge and vertex weights as

$$\mathbf{x}^{\mathsf{T}}\mathbf{L}\mathbf{x} = \sum_{i=1}^{n}(\mathbf{V})_{ii}\, x_i^2 + \sum_{(i,j)\in\mathcal{I}} (\mathbf{W})_{ij}\,(x_i - x_j)^2 \tag{20}$$

where $\mathbf{x} = [x_1, ..., x_n]^{\mathsf{T}}$, $(\mathbf{W})_{ij}=-(\mathbf{L})_{ij}$, $(\mathbf{V})_{ii}=\sum_{j=1}^{n}(\mathbf{L})_{ij}$ and $\mathcal{I}=\{(i,j)\,|\,(v_i,v_j)\in\mathcal{E}\}$ is the set of index pairs of vertices associated with the edge set $\mathcal{E}$. By inspecting (16) and (20), we can observe that assigning a larger (resp. smaller) edge weight (e.g., to $(\mathbf{W})_{ij}$) increases the probability of having a smaller (resp. larger) squared difference between corresponding data points (i.e., $x_i$ and $x_j$). Similarly, assigning a larger (resp. smaller) vertex weight (e.g., to $(\mathbf{V})_{ii}$) increases the probability of that the corresponding data point (i.e., $x_i$) will have smaller (resp. larger) magnitude. In a graph signal processing context [4], the term in (20) is also known as the *graph Laplacian quadratic form*, and is used to quantify smoothness of graph signals, with a smaller Laplacian quadratic term ($\mathbf{x}^{\mathsf{T}}\mathbf{L}\mathbf{x}$) indicating a smoother signal ($\mathbf{x}$). In our formulations, the Laplacian quadratic from $\mathbf{x}^{\mathsf{T}}\mathbf{L}\mathbf{x}$ relates to the trace term in our objective function, which is derived based on the likelihood function for GMRFs as discussed in the following.

The proposed graph learning problems can be probabilistically formulated as parameter estimation for attractive GMRFs from data. Assuming that $k$ independent, identically distributed samples, $\mathbf{x}_i$ for $i = 1, ..., k$, are obtained from an attractive GMRF with unknown parameters, the likelihood of a candidate graph Laplacian $\mathbf{L}$ can be written as

$$\prod_{i=1}^{k}\mathsf{p}(\mathbf{x}_i|\mathbf{L})=(2\pi)^{-\frac{kn}{2}}\,|\mathbf{L}^{\dagger}|^{-\frac{k}{2}}\prod_{i=1}^{k}\exp\left(-\frac{1}{2}{\mathbf{x}_i}^{\mathsf{T}}\mathbf{L}\mathbf{x}_i\right). \tag{21}$$

Let $\mathbf{L}(\mathbf{w},\mathbf{v})$ be defined by edge weight and vertex weight vectors $\mathbf{w} = [f_w(e_1), ..., f_w(e_m)]^{\mathsf{T}}$ and $\mathbf{v} = [f_v(v_1), ..., f_v(v_n)]^{\mathsf{T}}$, where $n$ is the number of vertices, and $m=n(n-1)/2$ is the number of all possible (undirected) edges. The maximization of the likelihood function in (21) can be equivalently formulated as minimizing the negative log-likelihood, that is

$$\begin{aligned}\widehat{\mathbf{L}}_{\mathrm{ML}} &= \underset{\mathbf{L}(\mathbf{w},\mathbf{v})}{\operatorname{argmin}}\left\{-\frac{k}{2}\log|\mathbf{L}|+\frac{1}{2}\sum_{i=1}^{k}\mathrm{Tr}\left({\mathbf{x}_i}^{\mathsf{T}}\mathbf{L}\mathbf{x}_i\right)\right\}\\ &= \underset{\mathbf{L}(\mathbf{w},\mathbf{v})}{\operatorname{argmin}}\left\{\mathrm{Tr}\left(\mathbf{L}\mathbf{S}\right)-\log|\mathbf{L}|\right\}\end{aligned} \tag{22}$$

where $\mathbf{S}$ is the sample covariance matrix, and $\widehat{\mathbf{L}}_{\mathrm{ML}}$ denotes the maximum likelihood estimate of $\mathbf{L}(\mathbf{w},\mathbf{v})$. Moreover, we can derive maximum a posteriori (MAP) estimation problems by incorporating the information known about $\mathbf{L}$ into a prior distribution $\mathsf{p}(\mathbf{L})$ as

$$\widehat{\mathbf{L}}_{\mathrm{MAP}} = \underset{\mathbf{L}(\mathbf{w},\mathbf{v})}{\operatorname{argmin}}\left\{\mathrm{Tr}\left(\mathbf{L}\mathbf{S}\right)-\log|\mathbf{L}|-\log(\mathsf{p}(\mathbf{L}))\right\}. \tag{23}$$

For example, we can choose the following $m$-variate exponential prior for sparse estimation of $\mathbf{w}$,

$$\mathsf{p}(\mathbf{w}) = (2\alpha)^m\exp\left(-2\alpha\mathbf{1}^{\mathsf{T}}\mathbf{w}\right) \quad \text{for } \mathbf{w}\geq\mathbf{0}, \tag{24}$$

so that the MAP estimation in (23) can be written as follows:

$$\begin{aligned}\widehat{\mathbf{L}}_{\mathrm{MAP}} &= \underset{\mathbf{L}(\mathbf{w},\mathbf{v})}{\operatorname{argmin}}\left\{\mathrm{Tr}\left(\mathbf{L}\mathbf{S}\right)-\log|\mathbf{L}|-\log(\mathsf{p}(\mathbf{w}))\right\}\\ &= \underset{\mathbf{L}(\mathbf{w},\mathbf{v})}{\operatorname{argmin}}\left\{\mathrm{Tr}\left(\mathbf{L}\mathbf{S}\right)-\log|\mathbf{L}|+2\alpha\|\mathbf{w}\|_1\right\}\\ &= \underset{\mathbf{L}(\mathbf{w},\mathbf{v})}{\operatorname{argmin}}\left\{\mathrm{Tr}\left(\mathbf{L}\mathbf{S}\right)-\log|\mathbf{L}|+\alpha\|\mathbf{L}\|_{1,\mathrm{off}}\right\}\end{aligned} \tag{25}$$

where the resulting minimization is equivalent to the objective of our problems with the regularization in (5).

**Proposition 5.** *Let the data model be* $\mathbf{x}\sim\mathsf{N}(\mathbf{0},\mathbf{L}^{\dagger})$ *as in (16). Then, Problems 1, 2 and 3 are specific instances of the maximum a posteriori estimation problem in (23).*

Hence, our graph learning problems can be interpreted as MAP parameter estimation for different classes of attractive GMRFs. Thus, when the data model assumptions are satisfied, solving our problems produces the optimal parameters in MAP sense. Given $\mathbf{S}$, which is obtained from the data, the solution of (23) corresponds to the closest Laplacian in terms of a regularized log-determinant divergence criterion [12]. In practice, in order to capture nonlinear relations between random variables, different types of kernels (e.g., polynomial and RBF kernels) can also be used to construct $\mathbf{S}$.

## V. Proposed Graph Learning Algorithms

Problems 1, 2 and 3 can be solved using general purpose solvers such as CVX [51]. However, these solvers generally implement second-order methods that require calculation of a Hessian matrix and are therefore computationally inefficient. Simpler gradient descent algorithms would also be computationally complex, since the full gradient calculation of the objective function involves inverting the current estimate of the Laplacian matrix at each iteration (e.g., see the derivative of (32) in (36)). In order to develop efficient methods, we propose iterative block-coordinate descent algorithms [47], where each iterate (block-coordinate update) is obtained by fixing some of the elements in the set of target variables while updating the rest. Thus, the original problem is decomposed into a series of lower-dimensional subproblems that are relatively easier to solve. Particularly, at each iteration, the update variables are formed by a row/column of the target graph Laplacian matrix ($\mathbf{\Theta}$), and they are updated by solving the subproblem derived based on the optimality conditions of corresponding Laplacian estimation problem, where the available structural constraints are also incorporated into the subproblem. Basically, to estimate an $n\times n$ graph Laplacian matrix, our algorithms iteratively update rows/columns of $\mathbf{\Theta}$ and its inverse ($\mathbf{C}$), so that the cycle of $n$ row/column updates is repeated until convergence is achieved. Also, depending on the type of target Laplacian, our algorithms potentially apply projections to satisfy the Laplacian constraints.

In what follows, we first provide matrix update formulas used to efficiently update entries of $\mathbf{\Theta}$ and $\mathbf{C}$ in our algorithms (Section V-A). Then, Algorithms 1 and 2 are presented with the derivations of subproblems based on the optimality conditions of corresponding graph learning problems[3]. Specifically, Algorithm 1 is proposed to solve Problems 1 and 2 (Section V-B), while Algorithm 2 solves Problem 3 (Section V-C). Finally, the convergence and computational complexity of proposed algorithms are analyzed (Section V-D).

### A. Matrix Update Formulas

The proposed algorithms exploit the following formulas to update the target variable $\mathbf{\Theta}$ and its inverse $\mathbf{C}$ iteratively.

**Row/column updates.** Updating the $u$-th row/column of $\mathbf{\Theta}$ results in updating all the elements in its inverse $\mathbf{C}=\mathbf{\Theta}^{-1}$, which can be obtained by the matrix inversion lemma [53],

$$(\mathbf{P}^{\mathsf{T}}\mathbf{\Theta}\mathbf{P})^{-1} = \begin{bmatrix} \mathbf{\Theta}_u & \boldsymbol{\theta}_u \\ \boldsymbol{\theta}_u^{\mathsf{T}} & \theta_u \end{bmatrix}^{-1} = \mathbf{P}^{\mathsf{T}}\mathbf{C}\mathbf{P} = \begin{bmatrix} \mathbf{C}_u & \mathbf{c}_u \\ \mathbf{c}_u^{\mathsf{T}} & c_u \end{bmatrix}$$
$$= \begin{bmatrix} \left(\mathbf{\Theta}_u - \dfrac{\boldsymbol{\theta}_u\boldsymbol{\theta}_u^{\mathsf{T}}}{\theta_u}\right)^{-1} & -\mathbf{C}_u\dfrac{\boldsymbol{\theta}_u}{\theta_u} \\ -\dfrac{\boldsymbol{\theta}_u^{\mathsf{T}}}{\theta_u}\mathbf{C}_u^{\mathsf{T}} & \dfrac{1}{\theta_u} - \dfrac{\boldsymbol{\theta}_u^{\mathsf{T}}\mathbf{C}_u\boldsymbol{\theta}_u}{\theta_u^2} \end{bmatrix} \quad (26)$$

where the permutation matrix $\mathbf{P}$ is used to arrange updated and fixed elements in block partitions, so that the submatrix $\mathbf{\Theta}_u$ represents the elements that remain unchanged, while vector $\boldsymbol{\theta}_u$ and scalar $\theta_u$ (i.e., $\theta_u=(\mathbf{\Theta})_{uu}$) are the $u$-th row/columns $\mathbf{\Theta}$, which are being updated. Based on the block partitions in (26), we can calculate $\mathbf{C}$, using updated $\boldsymbol{\theta}_u$ and $\theta_u$, for fixed $\mathbf{\Theta}_u$ as follows:

$$\mathbf{C}_u = \left(\mathbf{\Theta}_u - \frac{\boldsymbol{\theta}_u\boldsymbol{\theta}_u^{\mathsf{T}}}{\theta_u}\right)^{-1} = \mathbf{\Theta}_u^{-1} - \frac{\mathbf{\Theta}_u^{-1}\boldsymbol{\theta}_u\boldsymbol{\theta}_u^{\mathsf{T}}\mathbf{\Theta}_u^{-1}}{\theta_u - \boldsymbol{\theta}_u^{\mathsf{T}}\mathbf{\Theta}_u^{-1}\boldsymbol{\theta}_u}, \quad (27)$$

$$\mathbf{c}_u = -\mathbf{C}_u\frac{\boldsymbol{\theta}_u}{\theta_u} = -\frac{\mathbf{\Theta}_u^{-1}\boldsymbol{\theta}_u}{\theta_u - \boldsymbol{\theta}_u^{\mathsf{T}}\mathbf{\Theta}_u^{-1}\boldsymbol{\theta}_u}, \quad (28)$$

$$c_u = \frac{1}{\theta_u - \boldsymbol{\theta}_u^{\mathsf{T}}\mathbf{\Theta}_u^{-1}\boldsymbol{\theta}_u}, \quad (29)$$

where $\mathbf{\Theta}_u^{-1}$ can be calculated from partitions of updated $\mathbf{C}$ as,

$$\mathbf{\Theta}_u^{-1} = \mathbf{C}_u - \mathbf{c}_u\mathbf{c}_u^{\mathsf{T}}/c_u. \quad (30)$$

**Diagonal updates.** After adding a scalar value $\nu$ to $(\mathbf{\Theta})_{ii}$, we use the Sherman-Morrison formula [54] to update $\mathbf{C}$ as

$$\widehat{\mathbf{C}} = \widehat{\mathbf{\Theta}}^{-1} = (\mathbf{\Theta} + \nu\,\boldsymbol{\delta}_i\boldsymbol{\delta}_i^{\mathsf{T}})^{-1} = \mathbf{C} - \frac{\nu\,\mathbf{C}\boldsymbol{\delta}_i\boldsymbol{\delta}_i^{\mathsf{T}}\mathbf{C}}{1+\nu\boldsymbol{\delta}_i^{\mathsf{T}}\mathbf{C}\boldsymbol{\delta}_i}, \quad (31)$$

where $\boldsymbol{\delta}_i$ is the vector whose entries are zero, except for its $i$-th entry which is equal to one.

### B. Generalized Laplacian Estimation

**Derivation of the optimality conditions.** To derive necessary and sufficient optimality conditions, we use Lagrangian duality theory [55], [56], which requires introducing a set of Lagrange multipliers (i.e., dual variables) and a Lagrangian function. For Problems 1 and 2 the Lagrangian functions have the form,

$$-\mathrm{logdet}(\mathbf{\Theta}) + \mathrm{Tr}\,(\mathbf{\Theta}\mathbf{K}) + \mathrm{Tr}(\mathbf{\Theta}\mathbf{M}), \quad (32)$$

where $\mathbf{M}$ is the matrix of Lagrange multipliers associated with the constraints. In particular, for Problem 1, $\mathbf{M}=\mathbf{M}_1+\mathbf{M}_2$, with multiplier matrices, $\mathbf{M}_1$ and $\mathbf{M}_2$, whose entries are

$$(\mathbf{M}_1)_{ij} = (\mathbf{M}_1)_{ji} = \begin{cases} \mu_{ij}^{(1)} \geq 0 & \text{if } (\mathbf{A})_{ij}=1,\ i\neq j \\ 0 & \text{if } (\mathbf{A})_{ij}=0,\ i\neq j \\ 0 & \text{if } i=j \end{cases} \quad (33)$$

$$(\mathbf{M}_2)_{ij} = (\mathbf{M}_2)_{ji} = \begin{cases} \mu_{ij}^{(2)} \in \mathbb{R} & \text{if } (\mathbf{A})_{ij}=0,\ i\neq j \\ 0 & \text{if } (\mathbf{A})_{ij}=1,\ i\neq j \\ 0 & \text{if } i=j \end{cases} \quad (34)$$

for $i,j=1,...,n$ where $\mu_{ij}^{(1)}$ and $\mu_{ij}^{(2)}$ are the Lagrange multipliers associated with inequality and equality constraints in Problem 1, respectively. For Problem 2, $\mathbf{M}=\mathbf{M}_1+\mathbf{M}_2-\mathbf{M}_3$ so that $\mathbf{M}_1$ and $\mathbf{M}_2$ are as in (33) and (34), and $\mathbf{M}_3$ consists of Lagrange multipliers (denoted as $\mu_i^{(3)}$) associated with the constraint $\mathbf{\Theta}\mathbf{1} \geq \mathbf{0}$. The entries of $\mathbf{M}_3$ are

$$(\mathbf{M}_3)_{ij} = \mu_i^{(3)} + \mu_j^{(3)} \ \text{ where } \mu_i^{(3)} \geq 0,\ \mu_j^{(3)} \geq 0 \quad (35)$$

for $i,j=1,...,n$. By taking the derivative of (32) with respect to $\mathbf{\Theta}$ and setting it to zero, we obtain the following optimality condition,

$$-\mathbf{\Theta}^{-1} + \mathbf{K} + \mathbf{M} = \mathbf{O}, \quad (36)$$

and the necessary and sufficient optimality conditions [56] for Problem 1 are

$$\begin{aligned} &-\mathbf{\Theta}^{-1} + \mathbf{K} + \mathbf{M} = \mathbf{O} \\ &(\mathbf{\Theta})_{ij} \leq 0 \ \text{ if } (\mathbf{A})_{ij}=1,\ i\neq j \\ &(\mathbf{\Theta})_{ij} = 0 \ \text{ if } (\mathbf{A})_{ij}=0,\ i\neq j \\ &\mathbf{\Theta} \succeq 0 \quad (\mathbf{M}_1)_{ij}(\mathbf{\Theta})_{ij} = 0 \end{aligned} \quad (37)$$

where $\mathbf{M}=\mathbf{M}_1+\mathbf{M}_2$. For Problem 2, the multiplier matrix is $\mathbf{M}=\mathbf{M}_1+\mathbf{M}_2-\mathbf{M}_3$, and the corresponding optimality conditions include those in (37) as well as:

$$\mathbf{\Theta}\mathbf{1} \geq \mathbf{0} \quad (\mathbf{M}_3)_{ii}(\mathbf{\Theta}\mathbf{1})_i = 0. \quad (38)$$

**Subproblems for block-coordinate descent updates.** In our algorithm, we solve instances of the subproblem derived based on the optimality conditions of Problem 1. So, letting $\mathbf{C}=\mathbf{\Theta}^{-1}$ and using the conditions in (37), the optimality conditions for the $u$-th row/column of $\mathbf{\Theta}$ can be written as:

$$-\mathbf{c}_u + \mathbf{k}_u + \mathbf{m}_u = \mathbf{0} \quad (39)$$

$$-c_u + k_u = 0 \quad (40)$$

where the vectors, $\mathbf{c}_u$, $\mathbf{k}_u$ and $\mathbf{m}_u$, and the scalars, $c_u$, $k_u$ and $m_u$, are obtained by partitioning $\mathbf{C}$, $\mathbf{K}$ and $\mathbf{M}$, as in (26) such that $c_u=(\mathbf{C})_{uu}$, $k_u=(\mathbf{K})_{uu}$ and $m_u=(\mathbf{M})_{uu}=0$. By using the relations in (28) and (29), we can rewrite (39) as

$$\mathbf{\Theta}_u^{-1}\boldsymbol{\theta}_u c_u + \mathbf{k}_u + \mathbf{m}_u = \mathbf{0}. \quad (41)$$

Based on the relations in (37) and (40) the optimality conditions for the $u$-th column of $\mathbf{\Theta}$ (i.e., $\boldsymbol{\theta}_u$) include

$$\begin{aligned} &\mathbf{\Theta}_u^{-1}\boldsymbol{\theta}_u k_u + \mathbf{k}_u + \mathbf{m}_u = \mathbf{0} \\ &(\boldsymbol{\theta}_u)_i \leq 0 \ \text{ if } (\mathbf{a}_u)_i = 1 \\ &(\boldsymbol{\theta}_u)_i = 0 \ \text{ if } (\mathbf{a}_u)_i = 0 \end{aligned} \quad (42)$$

where $\boldsymbol{\theta}_u$ and $\mathbf{a}_u$ are obtained by partitioning $\mathbf{\Theta}$ and $\mathbf{A}$ as in (26), respectively, and the optimality conditions on $\mathbf{m}_u$ follow from (33) and (34). Based on the above optimality conditions, in (37) and (42), the optimal update for the $u$-th row/column of $\mathbf{\Theta}$ (i.e., $\boldsymbol{\theta}_u$) corresponds to the solution of the following quadratic program:

$$\begin{aligned} \underset{\boldsymbol{\theta}_u}{\text{minimize}} \quad & \frac{1}{2} k_u^2 \boldsymbol{\theta}_u^{\mathsf{T}} \mathbf{\Theta}_u^{-1} \boldsymbol{\theta}_u + k_u \boldsymbol{\theta}_u^{\mathsf{T}} \mathbf{k}_u \\ \text{subject to} \quad & (\boldsymbol{\theta}_u)_i \leq 0 \ \text{ if } (\mathbf{a}_u)_i = 1 \\ & (\boldsymbol{\theta}_u)_i = 0 \ \text{ if } (\mathbf{a}_u)_i = 0 \end{aligned} \tag{43}$$

[3]The implementations of our proposed algorithms are available in [52].

The above problem can be simplified by eliminating its equality constraints determined by $\mathbf{A}$ (i.e., $\mathbf{a}_u$), so that we formulate an equivalent version of (43) as the following nonnegative quadratic program [57], whose solution satisfies the optimality conditions in (42),

$$\begin{aligned} \underset{\boldsymbol{\beta}}{\text{minimize}} \quad & \frac{1}{2} \boldsymbol{\beta}^{\mathsf{T}} \mathbf{Q} \boldsymbol{\beta} - \boldsymbol{\beta}^{\mathsf{T}} \mathbf{p} \\ \text{subject to} \quad & \boldsymbol{\beta} \geq \mathbf{0} \end{aligned} \tag{44}$$

where

$$\begin{gathered} \boldsymbol{\beta} = -(\boldsymbol{\theta}_u)_{\mathcal{S}} \quad \mathbf{p} = (\mathbf{k}_u / k_u)_{\mathcal{S}} \quad \mathbf{Q} = (\mathbf{\Theta}_u^{-1})_{\mathcal{S}\mathcal{S}} \\ \mathcal{S} = \{ i \,|\, (\mathbf{a}_u)_i = 1 \} \end{gathered} \tag{45}$$

so that $\boldsymbol{\beta}$ is the vector whose elements are selected from the original variable vector $\boldsymbol{\theta}_u$ based on index set $\mathcal{S}$. For example, if $\mathcal{S} = \{1, 2, 5\}$, then $\boldsymbol{\beta} = -[(\boldsymbol{\theta}_u)_1, (\boldsymbol{\theta}_u)_2, (\boldsymbol{\theta}_u)_5]^{\mathsf{T}}$. Similarly, $\mathbf{Q}$ is constructed by selecting rows and columns of $\mathbf{\Theta}_u^{-1}$ with index values in $\mathcal{S}$, so the resulting $\mathbf{Q}$ is a submatrix of $\mathbf{\Theta}_u^{-1}$. It is important to note that the connectivity constraints (i.e., $\mathbf{A}$ or $\mathbf{a}_u$) allow us to reduce the dimension of the variable $\boldsymbol{\theta}_u$ and therefore, the dimension of (43).

For Problem 2, based on the conditions in (37) and (38), we can similarly formulate a quadratic program to update $u$-th row/column of $\mathbf{\Theta}$:

$$\begin{aligned} \underset{\boldsymbol{\theta}_u}{\text{minimize}} \quad & \frac{1}{2} c_u^2 \boldsymbol{\theta}_u^{\mathsf{T}} \mathbf{\Theta}_u^{-1} \boldsymbol{\theta}_u + c_u \boldsymbol{\theta}_u^{\mathsf{T}} \mathbf{k}_u \\ \text{subject to} \quad & (\boldsymbol{\theta}_u)_i \leq 0 \ \text{ if } (\mathbf{a}_u)_i = 1 \\ & (\boldsymbol{\theta}_u)_i = 0 \ \text{ if } (\mathbf{a}_u)_i = 0 \\ & -\boldsymbol{\theta}_u^{\mathsf{T}} \mathbf{1} \leq \theta_u \end{aligned} \tag{46}$$

where $\theta_u = (\mathbf{\Theta})_{uu}$. The above problem is also a nonnegative quadratic program. To solve (46) for all $u$, we first iteratively update each row/column of $\mathbf{\Theta}$ by solving the subproblem in (44). After completing a cycle of $n$ row/column updates, we modify the diagonal entries of the updated $\mathbf{\Theta}$, so that it satisfies the constraints in (46). The diagonal update parameters ($\boldsymbol{\nu}$) are the optimal solutions of the following projection problem for given $\mathbf{\Theta}$:

$$\begin{aligned} \underset{\boldsymbol{\nu}}{\text{minimize}} \quad & \| \mathbf{\Theta} - \widehat{\mathbf{\Theta}} \|_F^2 \\ \text{subject to} \quad & \widehat{\mathbf{\Theta}} = \mathbf{\Theta} + \operatorname{diag}(\boldsymbol{\nu}) \quad \widehat{\mathbf{\Theta}} \in \mathcal{L}_d \end{aligned} \tag{47}$$

where $\boldsymbol{\nu}$ is the vector of update parameters, and $\mathcal{L}_d$ denotes the set of diagonally dominant generalized Laplacian matrices.

**Proposed Algorithm.** Algorithm 1 is proposed to solve Problems 1 and 2 for a given connectivity matrix $\mathbf{A}$, type of desired Laplacian matrix (i.e., $\mathcal{L}_g$ or $\mathcal{L}_d$) and regularization matrix $\mathbf{H}$. Basically, the proposed algorithm iteratively updates each row/column of the working estimate of Laplacian matrix ($\widehat{\mathbf{\Theta}}$) and its inverse ($\widehat{\mathbf{C}}$) by solving the subproblem in (44). The main reason of updating $\mathbf{C}$ is that the derived subproblem is parametrized by $\mathbf{\Theta}_u^{-1}$, which depends on $\mathbf{C}$ as formulated in (30). In Algorithm 1, the for loop in lines 5–12 implements the cycle of $n$ row/column updates, where the update formulas (see lines 7, 9 and 10) are derived based on the relations in (27)–(30). If we are interested in solving Problem 1 (if $\mathcal{L} = \mathcal{L}_g$), then the algorithm skips the lines 13–19. For Problem 2 (for $\mathcal{L} = \mathcal{L}_d$) the for loop between the lines 14–18 iteratively modifies the diagonal elements of $\widehat{\mathbf{\Theta}}$ by solving the projection problem in (47) ensuring that the resulting $\widehat{\mathbf{\Theta}}$ is a diagonally dominant matrix. The inverse of $\widehat{\mathbf{\Theta}}$ (i.e., $\widehat{\mathbf{C}}$) is also iteratively updated, accordingly (see line 16). The overall procedure is repeated until a stopping criterion (line 20) has been satisfied.

**Algorithm 1** Generalized Graph Laplacian (GGL)

**Input:** Sample statistic $\mathbf{S}$, connectivity matrix $\mathbf{A}$, regularization matrix $\mathbf{H}$, target Laplacian set $\mathcal{L}$ and tolerance $\epsilon$
**Output:** $\mathbf{\Theta}$ and $\mathbf{C}$

1: Set $\mathbf{K} = \mathbf{S} + \mathbf{H}$
2: Initialize $\widehat{\mathbf{C}} = \operatorname{ddiag}(\mathbf{K})$ and $\widehat{\mathbf{\Theta}} = \widehat{\mathbf{C}}^{-1}$
3: **repeat**
4: Set $\widehat{\mathbf{\Theta}}_{\text{pre}} = \widehat{\mathbf{\Theta}}$
5: **for** $u = 1$ to $n$ **do**
6: Partition $\widehat{\mathbf{\Theta}}$, $\widehat{\mathbf{C}}$, $\mathbf{K}$ and $\mathbf{A}$ as in (26) for $u$
7: Update $\widehat{\mathbf{\Theta}}_u^{-1} = \widehat{\mathbf{C}}_u - \widehat{\mathbf{c}}_u \widehat{\mathbf{c}}_u^{\mathsf{T}} / \widehat{c}_u$
8: Solve (44) for $\boldsymbol{\beta}$ with $\mathbf{Q}$, $\mathbf{p}$ and $\mathcal{S}$ in (45)
9: Update $\widehat{\boldsymbol{\theta}}_u$ and $\widehat{\theta}_u$ using the solution $\widehat{\boldsymbol{\beta}}$ from above:
$$(\widehat{\boldsymbol{\theta}}_u)_{\mathcal{S}} = -\widehat{\boldsymbol{\beta}} \quad (\widehat{\boldsymbol{\theta}}_u)_{\mathcal{S}^c} = \mathbf{0}$$
$$\widehat{\theta}_u = 1/k_u + \widehat{\boldsymbol{\beta}}^{\mathsf{T}} \mathbf{Q} \widehat{\boldsymbol{\beta}}$$
10: Update $\widehat{\mathbf{c}}_u$, $\widehat{c}_u$ and $\widehat{\mathbf{C}}_u$:
$$\widehat{c}_u = 1/(\widehat{\theta}_u - \widehat{\boldsymbol{\theta}}_u^{\mathsf{T}} \widehat{\mathbf{\Theta}}_u^{-1} \widehat{\boldsymbol{\theta}}_u) \quad \widehat{\mathbf{c}}_u = \widehat{\mathbf{\Theta}}_u^{-1} \widehat{\boldsymbol{\theta}}_u / \widehat{c}_u$$
$$\widehat{\mathbf{C}}_u = \widehat{\mathbf{\Theta}}_u^{-1} + \widehat{\mathbf{c}}_u \widehat{\mathbf{c}}_u^{\mathsf{T}} / \widehat{c}_u$$
11: Rearrange $\widehat{\mathbf{\Theta}}$ and $\widehat{\mathbf{C}}$ using $\mathbf{P}$ for $u$ as in (26)
12: **end for**
13: **if** $\mathcal{L} = \mathcal{L}_d$ (target Laplacian is a DDGL) **then**
14: **for** $i = 1$ to $n$ **do**
15: **if** $(\widehat{\mathbf{\Theta}}\mathbf{1})_i < 0$ **then** $\nu = -(\widehat{\mathbf{\Theta}}\mathbf{1})_i$
16: Set $(\widehat{\mathbf{\Theta}})_{ii} = (\widehat{\mathbf{\Theta}})_{ii} + \nu$ and update $\widehat{\mathbf{C}}$ using (31)
17: **end if**
18: **end for**
19: **end if**
20: **until** criterion$(\widehat{\mathbf{\Theta}}, \widehat{\mathbf{\Theta}}_{\text{pre}}) \leq \epsilon$
21: **return** $\mathbf{\Theta} = \widehat{\mathbf{\Theta}}$ and $\mathbf{C} = \widehat{\mathbf{C}}$

### *C. Combinatorial Laplacian Estimation*

**Derivation of the optimality conditions.** Similar to our derivations in the previous subsection, in order to derive optimality conditions, we first define the Lagrangian function corresponding to Problem 3 as follows,

$$-\operatorname{logdet}(\mathbf{\Theta} + \mathbf{J}) + \operatorname{Tr}(\mathbf{\Theta}(\mathbf{K} + \mathbf{J})) + \operatorname{Tr}(\mathbf{\Theta}\mathbf{M}), \tag{48}$$

where $\mathbf{M} = \mathbf{M}_1 + \mathbf{M}_2 + \mathbf{M}_4$ consists of Lagrange multipliers associated with the constraints in (10) such that $\mathbf{M}_1$ and $\mathbf{M}_2$ are as defined in (33) and (34), and the entries of $\mathbf{M}_4$ are

$$(\mathbf{M}_4)_{ij} = \mu_i^{(4)} + \mu_j^{(4)} \ \text{ where } \mu_i^{(4)} \in \mathbb{R},\ \mu_j^{(4)} \in \mathbb{R} \tag{49}$$

for $i, j = 1, ..., n$. Based on the Lagrangian stated in (48), the necessary and sufficient optimality conditions for the problem

**Algorithm 2** Combinatorial Graph Laplacian (CGL)

**Input:** Sample statistic $\mathbf{S}$, connectivity matrix $\mathbf{A}$, regularization matrix $\mathbf{H}$ and tolerance $\epsilon$

**Output:** $\mathbf{\Theta}$ and $\mathbf{C}$

1: Set $\mathbf{J}=(1/n)\mathbf{1}\mathbf{1}^{\mathsf{T}}$ $\widetilde{\mathbf{K}}=\mathbf{S}+\mathbf{H}+\mathbf{J}$
2: Initialize $\widehat{\mathbf{C}}=\operatorname{ddiag}(\widetilde{\mathbf{K}})$ and $\widehat{\mathbf{\Theta}}=\widehat{\mathbf{C}}^{-1}$
3: **repeat**
4: Set $\widehat{\mathbf{\Theta}}_{\text{pre}}=\widehat{\mathbf{\Theta}}$
5: **for** $u=1$ to $n$ **do**
6: Partition $\widehat{\mathbf{\Theta}}$, $\widehat{\mathbf{C}}$, $\widetilde{\mathbf{K}}$ and $\mathbf{A}$ as in (26) for $u$
7: Calculate $\widehat{\mathbf{\Theta}}_u^{-1}=\widehat{\mathbf{C}}_u-\widehat{\mathbf{c}}_u\widehat{\mathbf{c}}_u^{\mathsf{T}}/\widehat{c}_u$
8: Solve (56) for $\boldsymbol{\beta}$ with $\mathbf{Q}$, $\mathbf{p}$ and $\mathcal{S}$ in (57)
9: Update $\widehat{\boldsymbol{\theta}}_u$ and $\widehat{\theta}_u$ using the solution $\widehat{\boldsymbol{\beta}}$ from above:

$$(\widehat{\boldsymbol{\theta}}_u)_{\mathcal{S}}=((1/n)\mathbf{1}-\widehat{\boldsymbol{\beta}})\quad(\widehat{\boldsymbol{\theta}}_u)_{\mathcal{S}^c}=(1/n)\mathbf{1}$$
$$\widehat{\theta}_u=1/\widetilde{k}_u+(\widehat{\boldsymbol{\beta}}-(1/n)\mathbf{1})^{\mathsf{T}}\mathbf{Q}(\widehat{\boldsymbol{\beta}}-(1/n)\mathbf{1})$$

10: Update $\widehat{\mathbf{c}}_u$, $\widehat{c}_u$ and $\widehat{\mathbf{C}}_u$:

$$\widehat{c}_u=1/(\widehat{\theta}_u-\widehat{\boldsymbol{\theta}}_u^{\mathsf{T}}\widehat{\mathbf{\Theta}}_u^{-1}\widehat{\boldsymbol{\theta}}_u)\quad\widehat{\mathbf{c}}_u=\widehat{\mathbf{\Theta}}_u^{-1}\widehat{\boldsymbol{\theta}}_u/\widehat{c}_u$$
$$\widehat{\mathbf{C}}_u=\widehat{\mathbf{\Theta}}_u^{-1}+\widehat{\mathbf{c}}_u\widehat{\mathbf{c}}_u^{\mathsf{T}}/\widehat{c}_u$$

11: Rearrange $\widehat{\mathbf{\Theta}}$ and $\widehat{\mathbf{C}}$ using $\mathbf{P}$ for $u$ as in (26)
12: **end for**
13: **for** $i=1$ to $n$ **do**
14: **if** $(\widehat{\mathbf{\Theta}}\mathbf{1})_i-1\neq 0$ **then** $\nu=-(\widehat{\mathbf{\Theta}}\mathbf{1})_i+1$
15: Set $(\widehat{\mathbf{\Theta}})_{ii}=(\widehat{\mathbf{\Theta}})_{ii}+\nu$ and update $\widehat{\mathbf{C}}$ using (31)
16: **end if**
17: **end for**
18: **until** criterion$(\widehat{\mathbf{\Theta}},\widehat{\mathbf{\Theta}}_{\text{pre}})\leq\epsilon$
19: **return** $\mathbf{\Theta}=\widehat{\mathbf{\Theta}}-\mathbf{J}$ and $\mathbf{C}=\widehat{\mathbf{C}}-\mathbf{J}$

in (10) are

$$\begin{aligned}&-\widetilde{\mathbf{\Theta}}^{-1}+\widetilde{\mathbf{K}}+\mathbf{M}_1+\mathbf{M}_2+\mathbf{M}_4=\mathbf{O}\\&(\widetilde{\mathbf{\Theta}})_{ij}\leq 1/n\ \text{ if } (\mathbf{A})_{ij}=1,\ i\neq j\\&(\widetilde{\mathbf{\Theta}})_{ij}=1/n\ \text{ if } (\mathbf{A})_{ij}=0,\ i\neq j\\&\widetilde{\mathbf{\Theta}}\succeq 0\quad\widetilde{\mathbf{\Theta}}\mathbf{1}=\mathbf{1}\quad(\mathbf{M}_1)_{ij}\left((\widetilde{\mathbf{\Theta}})_{ij}-1/n\right)=0\end{aligned}\tag{50}$$

where $\widetilde{\mathbf{\Theta}}=\mathbf{\Theta}+\mathbf{J}$, $\widetilde{\mathbf{C}}=(\mathbf{\Theta}+\mathbf{J})^{-1}$ and $\widetilde{\mathbf{K}}=\mathbf{K}+\mathbf{J}$. The matrices $\mathbf{M}_1$, $\mathbf{M}_2$ and $\mathbf{M}_4$ are defined as in (33), (34) and (49), respectively. For the $u$-th row/column of $\widetilde{\mathbf{\Theta}}$, the first optimality condition in (50) reduces to

$$-\widetilde{\mathbf{c}}_u+\widetilde{\mathbf{k}}_u+\mathbf{m}_u=\mathbf{0}\tag{51}$$

$$-\widetilde{c}_u+\widetilde{k}_u+m_u=0\tag{52}$$

where the condition in (51) can also be stated using the relations in (28) and (29) as

$$\widetilde{\mathbf{\Theta}}_u^{-1}\widetilde{\boldsymbol{\theta}}_u\widetilde{c}_u+\widetilde{\mathbf{k}}_u+\mathbf{m}_u=\mathbf{0}.\tag{53}$$

**Subproblem for block-coordinate descent updates.** Based on the optimality conditions stated in (50) and (53), we derive the following quadratic program solved for updating $u$-th row/column of $\widetilde{\mathbf{\Theta}}$,

$$\begin{aligned}\underset{\widetilde{\boldsymbol{\theta}}_u}{\text{minimize}}\quad&\frac{1}{2}\widetilde{c}_u^2\widetilde{\boldsymbol{\theta}}_u^{\mathsf{T}}\widetilde{\mathbf{\Theta}}_u^{-1}\widetilde{\boldsymbol{\theta}}_u+\widetilde{c}_u\widetilde{\boldsymbol{\theta}}_u^{\mathsf{T}}\widetilde{\mathbf{k}}_u\\\text{subject to}\quad&(\widetilde{\boldsymbol{\theta}}_u)_i\leq 1/n\ \text{ if } (\mathbf{a}_u)_i=1\\&(\widetilde{\boldsymbol{\theta}}_u)_i=1/n\ \text{ if } (\mathbf{a}_u)_i=0\\&-(\widetilde{\boldsymbol{\theta}}_u-(1/n)\mathbf{1})^{\mathsf{T}}\mathbf{1}=\widetilde{\theta}_u-(1/n)\end{aligned}\tag{54}$$

By changing variables $\widetilde{\boldsymbol{\theta}}_u=\boldsymbol{\theta}_u+(1/n)\mathbf{1}$, $\widetilde{\theta}_u=\theta_u+(1/n)$ and dividing the objective function with $\widetilde{c}_u^2$, we rewrite (54) as a quadratic program of the standard form,

$$\begin{aligned}\underset{\boldsymbol{\theta}_u}{\text{minimize}}\quad&\frac{1}{2}\boldsymbol{\theta}_u^{\mathsf{T}}\widetilde{\mathbf{\Theta}}_u^{-1}\boldsymbol{\theta}_u+\boldsymbol{\theta}_u^{\mathsf{T}}\left(\frac{\widetilde{\mathbf{k}}_u}{\widetilde{c}_u}+\frac{1}{n}\widetilde{\mathbf{\Theta}}_u^{-1}\mathbf{1}\right)\\\text{subject to}\quad&(\boldsymbol{\theta}_u)_i\leq 0\ \text{ if } (\mathbf{a}_u)_i=1\\&(\boldsymbol{\theta}_u)_i=0\ \text{ if } (\mathbf{a}_u)_i=0\\&-\boldsymbol{\theta}_u^{\mathsf{T}}\mathbf{1}=\theta_u\end{aligned}\tag{55}$$

which can be simplified by eliminating the equality constraints as follows,

$$\begin{aligned}\underset{\boldsymbol{\beta}}{\text{minimize}}\quad&\frac{1}{2}\boldsymbol{\beta}^{\mathsf{T}}\mathbf{Q}\boldsymbol{\beta}-\boldsymbol{\beta}^{\mathsf{T}}\mathbf{p}\\\text{subject to}\quad&\boldsymbol{\beta}\geq\mathbf{0}\end{aligned}\tag{56}$$

where

$$\begin{aligned}\boldsymbol{\beta}=-(\boldsymbol{\theta}_u)_{\mathcal{S}}\quad&\mathbf{p}=(\widetilde{\mathbf{k}}_u/\widetilde{k}_u+(1/n)\widetilde{\mathbf{\Theta}}_u^{-1}\mathbf{1})_{\mathcal{S}}\\\mathbf{Q}=(\widetilde{\mathbf{\Theta}}_u^{-1})_{\mathcal{S}\mathcal{S}}\quad&\mathcal{S}=\{i\,|\,(\mathbf{a}_u)_i=1\}.\end{aligned}\tag{57}$$

In order to solve (55) for all $u$, we first iteratively update each row/column by solving the nonnegative quadratic program in (56). After each cycle of $n$ row/column updates, the diagonal entries of the resulting matrix $(\widehat{\mathbf{\Theta}})$ are modified to satisfy the combinatorial Laplacian constraints $-\boldsymbol{\theta}_u^{\mathsf{T}}\mathbf{1}=\theta_u$ for $u=1,...,n$ in (55). The diagonal update parameters are optimized by solving the following projection problem for given $\widehat{\mathbf{\Theta}}$

$$\begin{aligned}\underset{\boldsymbol{\nu}}{\text{minimize}}\quad&\|\widetilde{\mathbf{\Theta}}-\widehat{\mathbf{\Theta}}\|_F^2\\\text{subject to}\quad&\widetilde{\mathbf{\Theta}}=\widehat{\mathbf{\Theta}}+\operatorname{diag}(\boldsymbol{\nu})\quad(\widetilde{\mathbf{\Theta}}-\mathbf{J})\in\mathcal{L}_c\end{aligned}\tag{58}$$

where $\boldsymbol{\nu}$ is the vector of update parameters, $\mathcal{L}_c$ denotes the set of combinatorial Laplacian matrices and $\mathbf{J}=(1/n)\mathbf{1}\mathbf{1}^{\mathsf{T}}$.

**Proposed Algorithm.** Algorithm 2 is proposed to solve Problem 3 for a given connectivity matrix $\mathbf{A}$ and regularization matrix $\mathbf{H}$. Although the basic structures of Algorithms 1 and 2 are similar, Algorithm 2 has three major differences. Firstly, the algorithm does not directly estimate the target Laplacian matrix (i.e., $\mathbf{\Theta}$). Instead, it iteratively solves for the matrix $\widetilde{\mathbf{\Theta}}=\mathbf{\Theta}+\mathbf{J}$ whose entries are shifted by $(1/n)$. Secondly, the subproblem solved for updating each row/column and the associated update formulas are different (see lines 8, 9 and 10 in Algorithm 2). Thirdly, the for loop in lines 13–17 maintains that the estimate of $\widetilde{\mathbf{\Theta}}$ leads to a CGL matrix (via the $\widehat{\mathbf{\Theta}}-\mathbf{J}$ transformation) by solving the projection problem in (58).

In Algorithm 2, we propose to estimate the shifted matrix $\widetilde{\mathbf{\Theta}}$ because of the singularity of combinatorial Laplacian matrices, by their construction. However, our result in Proposition 1 shows that one can solve the CGL problem in (9) by solving the equivalent problem in (10). Based on this result, we derive optimality conditions for (10), and develop Algorithm 2 which iteratively estimates $\widetilde{\mathbf{\Theta}}$ until the convergence criterion has been satisfied. Then, the optimal $\mathbf{\Theta}$ is recovered by subtracting $\mathbf{J}$ from the estimated $\widetilde{\mathbf{\Theta}}$ as in line 19 of Algorithm 2.

### *D. Convergence and Complexity Analysis of Algorithms*

**Convergence.** The following proposition addresses the convergence of our proposed algorithms based on the results from [47] [55] [58] (see Appendix for the proof).

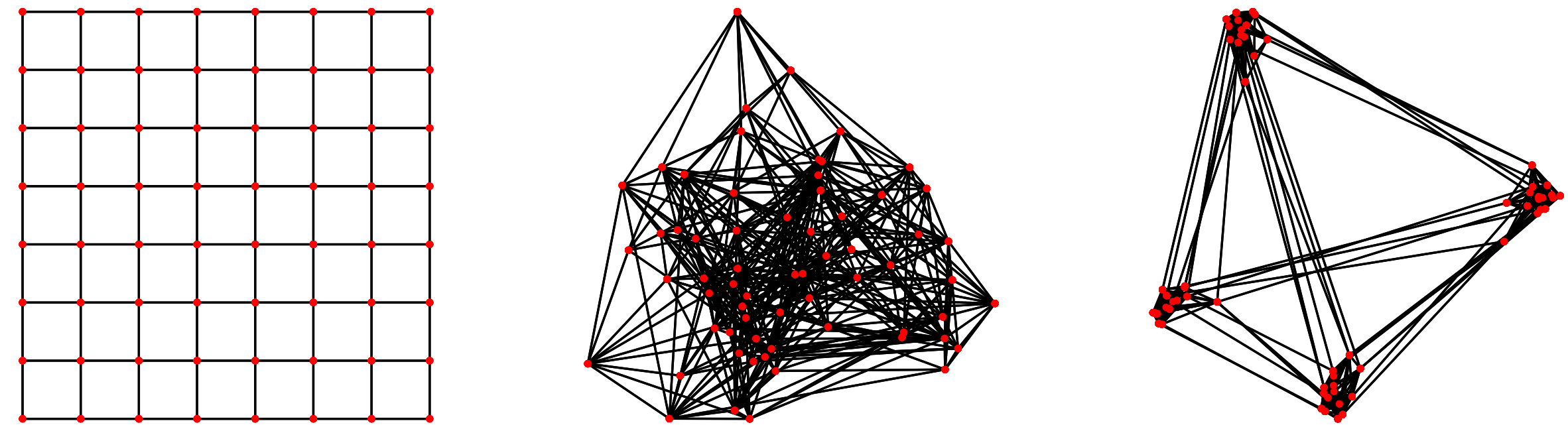

(a) $\mathcal{G}_{\text{grid}}^{(64)}$: 8×8 grid graph (b) $\mathcal{G}_{\text{ER}}^{(64,0.2)}$: Erdos-Renyi graph (c) $\mathcal{G}_{\text{M}}^{(64,0.2,0.9)}$: Modular graph (4 modules)

Fig. 2. Examples of different graph connectivity models, (a) grid (b) Erdos-Renyi (c) modular graphs used in our experiments. All graphs have 64 vertices.

**Proposition 6.** *Algorithms 1 and 2 guarantee convergence to the global optimal solutions of Problems 1, 2 and 3.*

**Complexity.** In Algorithms 1 and 2, each block-coordinate descent iteration has $O(T_p(n)+n^2)$ complexity where $O(T_p(n))$ is the worst-case complexity of solving the subproblem with dimension $n$, and the $n^2$ term is due to updating $\widehat{\mathbf{\Theta}}_u^{-1}$. After a cycle of $n$ row/column updates, updating a diagonal entry of $\widehat{\mathbf{\Theta}}$ and its inverse, $\widehat{\mathbf{C}}$, also has $O(n^2)$ complexity. Depending on the sparsity of solutions (i.e., graphs) the complexity can be reduced to $O(T_p(s)+n^2)$ where $s$ is the maximum number of edges connected to a vertex (i.e., number of non-zero elements in any row/column of $\mathbf{\Theta}$). Moreover, both proposed algorithms use diagonal matrices to initialize $\widehat{\mathbf{\Theta}}$ and $\widehat{\mathbf{C}}$. In practice, better initializations (i.e., "warm starts") and randomized implementations [47] can be exploited to reduce the algorithm runtime. To solve the subproblems in (44) and (56), which are specifically nonnegative quadratic programs, we employ an extension of Lawson-Hanson algorithm [59] with a block principal pivoting method [60]. Since nonnegative quadratic programs require varying number of iterations to converge for each row/column update, it is hard to characterize the overall complexity of our algorithms. Yet, the complexity of solving the subproblems is $T_p(n)=\Omega(n^3)$, which can be significantly reduced if the solutions are sparse ($T_p(s)=\Omega(s^3)$ for $s$-sparse solutions). In Section VI, we present empirical complexity results for the proposed algorithms.

## VI. Experimental Results

In this section, we present experimental results for comprehensive evaluation of our proposed graph learning techniques. Firstly, we compare the accuracy of our proposed algorithms and the state-of-the-art methods, where the datasets are artificially generated based on attractive GMRFs. Specifically, our GGL and DDGL estimation methods based on Algorithm 1 (GGL and DDGL) are evaluated by benchmarking against the Graphical Lasso algorithm (GLasso) [33]. Algorithm 2 (CGL) is compared to the methods proposed for estimating shifted CGL matrices (SCGL) [41] and learning graphs from smooth signals (GLS) [42][43], as well as the graph topology inference approach (GTI) proposed in [44]. Secondly, empirical computational complexity results are presented, and the advantage of exploiting connectivity constraints is demonstrated. Thirdly, the proposed methods are applied to learn similarity graphs from a real categorical dataset with binary entries, and the results are visually investigated. Finally, we evaluate the effect of model mismatch and of incorporating inaccurate connectivity constraints on graph Laplacian estimation.

### A. Comparison of Graph Learning Methods

**Datasets.** In order to demonstrate the performance of the proposed algorithms, we create several artificial datasets based on different graph-based models. Then the baseline and proposed algorithms are used to recover graphs (i.e., graph-based models) from the artificially generated data. To create a dataset, we first construct a graph, then its associated Laplacian matrix, $\mathbf{L}$, is used to generate independent data samples from the distribution $\mathsf{N}(\mathbf{0},\mathbf{L}^\dagger)$. A graph (i.e., $\mathbf{L}$) is constructed in two steps. In the first step, we determine the graph structure (i.e., connectivity) based on one of the following three options (see Fig. 2):

- Grid graph, $\mathcal{G}_{\text{grid}}^{(n)}$, consisting of $n$ vertices attached to their four nearest neighbors (except the vertices at boundaries).
- Random Erdos-Renyi graph, $\mathcal{G}_{\text{ER}}^{(n,p)}$, with $n$ vertices attached to other vertices with probability $p$.
- Random modular graph (also known as stochastic block model), $\mathcal{G}_{\text{M}}^{(n,p_1,p_2)}$ with $n$ vertices and four modules (subgraphs) where the vertex attachment probabilities across modules and within modules are $p_1$ and $p_2$, respectively.

In the second step, the graph weights (i.e., edge and vertex weights) are randomly selected based on a uniform distribution from the interval $[0.1, 3]$, denoted as $\mathsf{U}(0.1, 3)$. Note that this procedure leads to DDGLs, which are used in comparing our Algorithm 1 against GLasso. For evaluation of Algorithm 2, the edge weights are randomly selected from the same distribution $\mathsf{U}(0.1, 3)$, while all vertex weights are set to zero.

**Experimental setup.** For comprehensive evaluation, we create various experimental scenarios by choosing different graph structures (grid, Erdos-Renyi or modular) with varying number of vertices. Particularly, the GGL, DDGL and GLasso methods are tested on graphs consisting of 64 or 256 vertices. Since SCGL and GTI approaches [41] [44] do not currently have an efficient algorithm, the CGL estimation methods are only evaluated on graphs with 36 vertices. We also test the performance of the proposed methods with and without connectivity constraints. For each scenario, Monte-Carlo simulations are performed to test baseline and proposed algorithms with varying number of data samples ($k$). All algorithms use the following convergence criterion with the tolerance $\epsilon = 10^{-4}$,

$$\mathsf{criterion}(\widehat{\mathbf{\Theta}}, \widehat{\mathbf{\Theta}}_{\text{pre}}) = \frac{\|\widehat{\mathbf{\Theta}} - \widehat{\mathbf{\Theta}}_{\text{pre}}\|_F}{\|\widehat{\mathbf{\Theta}}_{\text{pre}}\|_F} \leq \epsilon \,, \tag{59}$$

where $\widehat{\mathbf{\Theta}}$ and $\widehat{\mathbf{\Theta}}_{\text{pre}}$ denote the graph parameters from current and previous steps, respectively (see Algorithms in 1 and 2).

In order to measure graph learning performance, we use two different metrics,

$$\mathsf{RE}(\widehat{\mathbf{\Theta}}, \mathbf{\Theta}^*) = \frac{\|\widehat{\mathbf{\Theta}} - \mathbf{\Theta}^*\|_F}{\|\mathbf{\Theta}^*\|_F} \tag{60}$$

which is the relative error between the ground truth graph ($\mathbf{\Theta}^*$) and estimated graph parameters ($\widehat{\mathbf{\Theta}}$), and

$$\mathsf{FS}(\widehat{\mathbf{\Theta}}, \mathbf{\Theta}^*) = \frac{2\,\mathsf{tp}}{2\,\mathsf{tp} + \mathsf{fn} + \mathsf{fp}} \tag{61}$$

is the F-score metric (commonly used metric to evaluate binary classification performance) calculated based on true-positive (tp), false-positive (fp) and false-negative (fn) detection of graph edges in estimated $\widehat{\mathbf{\Theta}}$ with respect to the ground truth edges in $\mathbf{\Theta}^*$. F-score takes values between 0 and 1, where the value 1 means perfect classification.

In our experiments, since SCGL and GLasso methods employ $\alpha\|\mathbf{\Theta}\|_1$ for regularization, we use the same regularization in our proposed methods (i.e., the matrix $\mathbf{H}$ is selected as in (4) for Problems 1, 2 and 3) to fairly compare all methods. Monte-Carlo simulations are performed for each proposed/baseline method, by successively solving the associated graph learning problem with different regularization parameters (i.e., $\alpha$, $\alpha_1$ and $\alpha_2$) to find the (best) regularization minimizing RE. The corresponding graph estimate is also used to calculate FS. Particularly, for the GGL, DDGL, CGL, SCGL and GLasso methods, $\alpha$ is selected from the following set:

$$\{0\} \cup \{0.75^r (s_{\text{max}} \sqrt{\log(n)/k}) \,|\, r = 1, 2, 3, \cdots, 14\}, \tag{62}$$

where $s_{\text{max}} = \max_{i \neq j} |(\mathbf{S})_{ij}|$ is the maximum off-diagonal entry of $\mathbf{S}$ in absolute value, and the scaling term $\sqrt{\log(n)/k}$ is used for adjusting the regularization according to $k$ and $n$ as suggested in [61] [62]. For both of the GLS methods [42] [43] addressing the problems in (13) and (14), $\alpha_1$ is selected from $\{0\} \cup \{0.75^r s_{\text{max}} \,|\, r = 1, 2, 3, \cdots, 14\}$, and for the problem in (14) the parameter $\alpha_2$ is selected by fine tuning. For GLS [42], the $\text{Tr}(\mathbf{\Theta}) = n$ constraint in (13) is set as $\text{Tr}(\mathbf{\Theta}) = \text{Tr}(\mathbf{\Theta}^*)$. Since GLS [43] and GTI [44] approaches generally result in severely biased solutions with respect to the ground truth $\mathbf{\Theta}^*$ (based on our observations from the experiments), their RE values are calculated after normalizing the estimated solution $\mathbf{\Theta}$ as $\widehat{\mathbf{\Theta}} = (\text{Tr}(\mathbf{\Theta}^*)/\text{Tr}(\mathbf{\Theta}))\mathbf{\Theta}$.

**Results.** Figure 3 demonstrates graph learning performances of different methods (in terms of average RE and FS) with respect to the number of data samples, used to calculate sample covariance $\mathbf{S}$, per number of vertices (i.e., $k/n$). In our results, GGL$(\mathbf{A}, \alpha)$, DDGL$(\mathbf{A}, \alpha)$ and CGL$(\mathbf{A}, \alpha)$ refer to solving the graph learning problems with both connectivity constraints and regularization, where the constraints are determined based on the true graph connectivity. GGL$(\alpha)$, DDGL$(\alpha)$ and CGL$(\alpha)$ refer to solving the problems with $\ell_1$-regularization only (i.e., without connectivity constraints).

TABLE II
AVERAGE RELATIVE ERRORS FOR DIFFERENT GRAPH CONNECTIVITY MODELS AND NUMBER OF VERTICES WITH FIXED $k/n = 30$

| Connectivity models | Average RE$(n=64)$ \| Average RE$(n=256)$ | | |
|---|---|---|---|
| | GLasso$(\alpha)$ | GGL$(\alpha)$ | GGL$(\mathbf{A}, \alpha)$ |
| $\mathcal{G}^{(n)}_{\text{grid}}$ | 0.079 \| 0.078 | 0.053 \| 0.037 | 0.040 \| 0.027 |
| $\mathcal{G}^{(n,0.1)}_{\text{ER}}$ | 0.105 \| 0.112 | 0.077 \| 0.082 | 0.053 \| 0.053 |
| $\mathcal{G}^{(n,0.1,0.3)}_{\text{M}}$ | 0.102 \| 0.124 | 0.075 \| 0.081 | 0.051 \| 0.053 |

As shown in Fig. 3, our proposed methods outperform all baseline approaches (namely GLasso [33], SCGL [41], GLS [42] [43] and GTI [44]) in terms of both RE and FS metrics. Naturally, incorporating the connectivity constraints (e.g., in GGL$(\mathbf{A}, \alpha)$ and CGL$(\mathbf{A}, \alpha)$) significantly improves the graph learning performance. However, for small $k/n$, it may not provide a perfect FS, even if the true graph connectivity is given. This is because there may not be a sufficient number of data samples to effectively recover the graph information. Specifically for $k/n \leq 1$, both $\mathbf{S}$ and the estimated graph Laplacian have low-rank. Since low-rank graph Laplacians correspond to disconnected graphs (i.e., graphs with more than one connected components), they can cause false-negative (fn) detections which reduce FS. Furthermore, the proposed methods outperform all baseline approaches regardless of the size of the graph ($n$) and the connectivity model ($\mathbf{A}$). As an example, for fixed $k/n = 30$, Table II compares average RE results for different graph connectivity models and number of vertices ($n$ for 64 and 256). Also, Figs. 4 and 5 illustrate sample GGL and CGL estimation results and compare different methods.

For estimation of GGL matrices, Fig. 3a demonstrates that the best set of results are obtained by solving the DDGL problem, DDGL$(\mathbf{A}, \alpha)$ and DDGL$(\alpha)$. This is expected since the random data samples are generated based on DDGL matrices (as part of our experimental setup), and exploiting additional information about the type of graph Laplacian improves graph learning performance. Generally, solving the GGL problem, GGL$(\mathbf{A}, \alpha)$ and GGL$(\alpha)$, also provides a good performance, where the difference between GGL and DDGL is often negligible. Moreover, both of the proposed GGL$(\alpha)$ and DDGL$(\alpha)$ methods (i.e., Algorithm 1) perform considerably better than GLasso, especially when the number of data samples ($k/n$) is small, since exploiting Laplacian constraints fulfills the model assumptions of attractive GMRFs. For estimation of CGL matrices, Fig. 3b shows that CGL (i.e., Algorithm 2) provides significantly better performance with increasing number of data samples ($k/n$) as compared to the SCGL, GLS and GTI approaches. Particularly, SCGL and GLS have limited accuracy even for large number of samples (e.g., $k/n \geq 100$). The main reason is due to their problem formulations, where SCGL [41] optimizes a shifted CGL instead of directly estimating a CGL matrix as stated in (12). The GLS and GTI methods solve the problems in (13), (14) and (15), whose objective functions

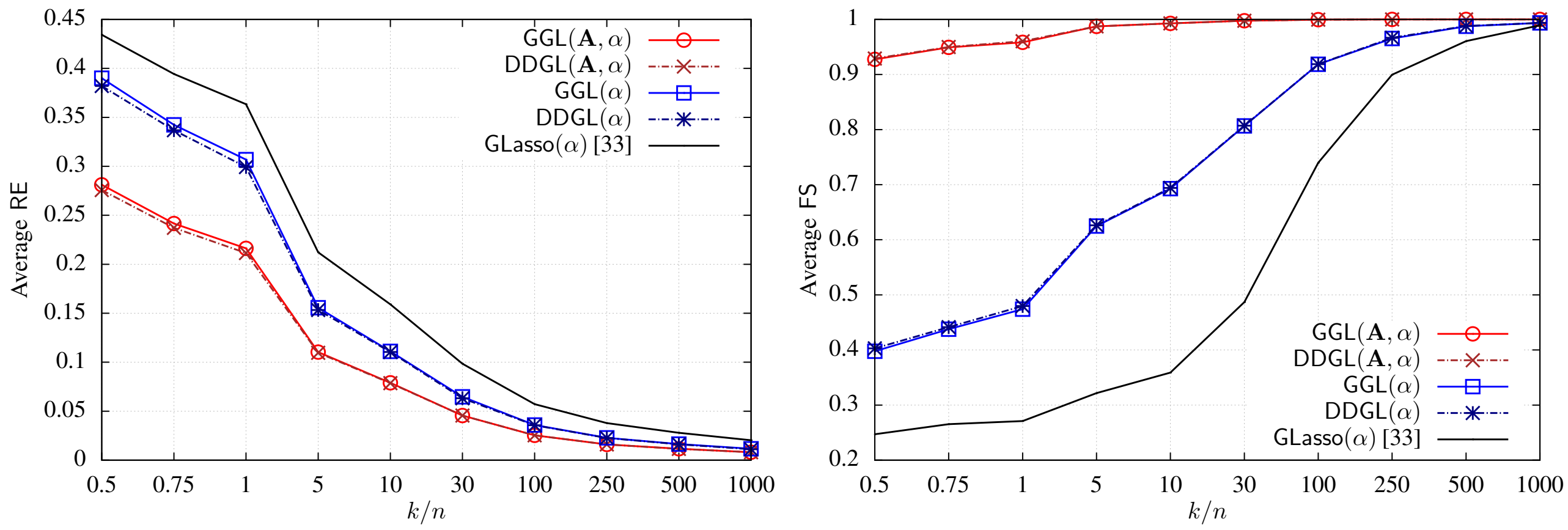

(a) Average performance results for learning generalized graph Laplacian matrices: The proposed methods GGL$(\alpha)$ and DDGL$(\alpha)$ outperform GLasso$(\alpha)$. The degree of improvement achieved by GGL$(\mathbf{A}, \alpha)$ and DDGL$(\mathbf{A}, \alpha)$, incorporating the connectivity constraints, is also demonstrated.

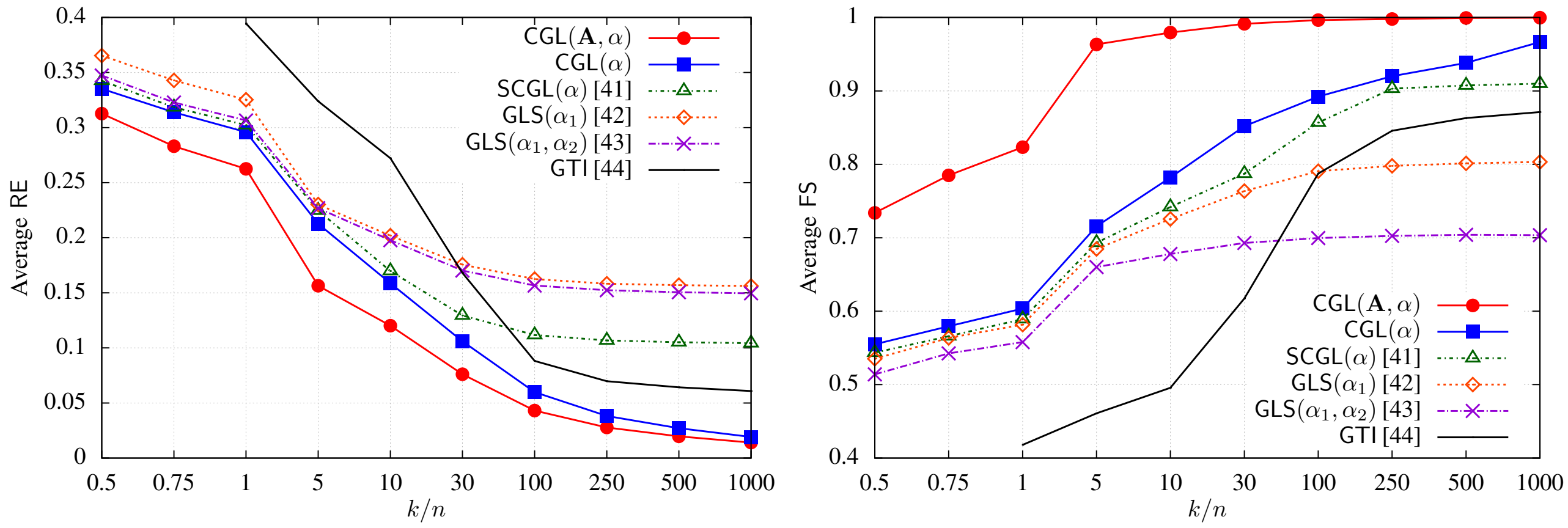

(b) Average performance results for learning combinatorial graph Laplacian matrices: The proposed CGL$(\alpha)$ method outperforms all baseline approaches, and the difference increases as gets $k/n$ larger. Incorporating the connectivity constraints (i.e., CGL$(\mathbf{A}, \alpha)$) naturally improves the performance.

Fig. 3. Comparison of the (a) GGL and (b) CGL estimation methods: The algorithms are tested on grid ($\mathcal{G}^{(n)}_{\text{grid}}$) and random graphs ($\mathcal{G}^{(n,0.1)}_{\text{ER}}$ and $\mathcal{G}^{(n,0.1,0.3)}_{\text{M}}$).

are derived without taking multivariate Gaussian distributions into account. Specifically, the objective in (14) serves as a lower-bound for the maximum-likelihood criterion in (22) (see Proposition 4). Besides, the performance difference against GTI is substantial [44] across different $k/n$, and GTI generally does not converge if $\mathbf{S}$ has low-rank (i.e., $k/n < 1$), so those results are not available.

### *B. Empirical Results for Computational Complexity*

Figure 7 compares the computational speedups achieved by our proposed algorithms over GLasso, which is implemented according to the P-GLasso algorithm presented in [34]. In our experiments, the algorithms are tested on Erdos-Renyi graphs with $n = 64$ vertices ($\mathcal{G}^{(64,p)}_{\text{ER}}$). By varying the parameter $p$, we evaluate the speedups at different graph sparsity levels ($p = 1$ means a fully connected graph). For each $p$, 10 different graphs and associated datasets (with $k/n = 1000$) are generated as discussed in the previous section. The speedup values are calculated using $\bar{T}_{\text{GLasso}}/\bar{T}$, where $\bar{T}$ and $\bar{T}_{\text{GLasso}}$ denote average execution times of the test algorithm (Algorithm 1 or 2) and GLasso algorithm, respectively. Since GLasso is approximately 1.5 times faster than both GLS methods [42] [43] on average, and the other methods [41] [44] do not have efficient implementations, we only use GLasso as the baseline algorithm in this experiment.

As shown in Fig. 7, the proposed methods provide faster convergence than GLasso regardless of $p$ (i.e., sparsity level), and the computational gain is substantial for learning sparse graphs (e.g., $p \leq 0.3$), where incorporating the connectivity constraints contributes to an additional 2 to 3-fold speedup over the methods without exploiting connectivity constraints. In the worst case, for dense graphs (e.g., $p \geq 0.8$), our methods converge approximately 5 times faster than the GLasso. This is mainly because, at each iteration, GLasso solves an $\ell_1$-regularized quadratic program [33] [34] having a nonsmooth objective function, and it is generally harder to solve compared to the (smooth) nonnegative quadratic program in (44).

### *C. Illustrative Results on Real Data*

We present some illustrative results to demonstrate that the proposed methods can also be useful to represent categorical (non-Gaussian) data. In our experiments, we use the *Animals dataset* [63] to learn weighted graphs, where graph vertices denote animals and edge weights represent the similarity

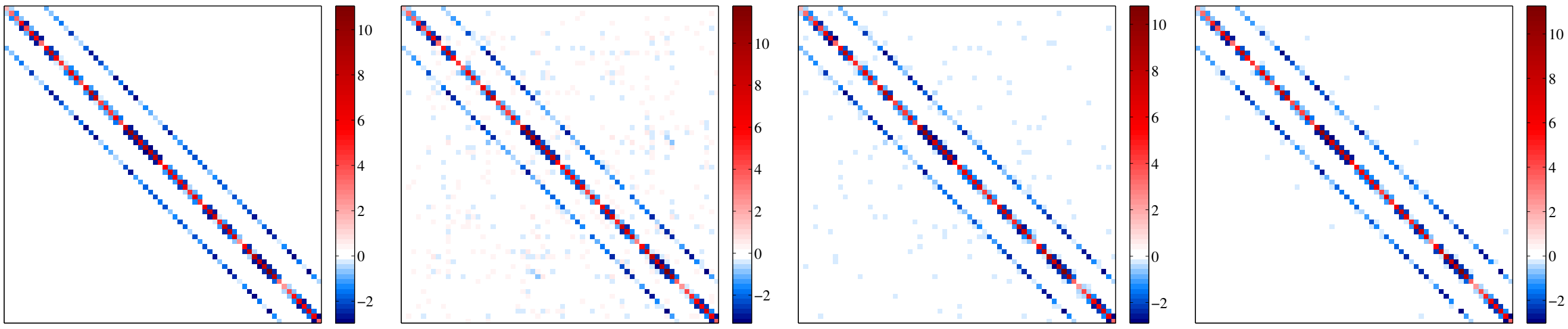

(a) The ground truth GGL ($\boldsymbol{\Theta}^*$) (b) $\mathsf{S}^{-1}$: (RE,FS) = (0.171,0.20) (c) GLasso: (RE,FS) = (0.078,0.64) (d) GGL: (RE,FS) = (0.054,0.89)

Fig. 4. Illustration of precision matrices (whose entries are color coded) estimated from $\mathbf{S}$ for $k/n = 30$: In this example, (a) the ground truth is a GGL associated with a grid graph having $n = 64$ vertices. From $\mathbf{S}$, the matrices are estimated by (b) inverting $\mathbf{S}$, (c) GLasso($\alpha$) and (d) GGL($\alpha$). The proposed method provides the best estimation in terms of RE and FS, and the resulting matrix is visually the most similar to the ground truth $\boldsymbol{\Theta}^*$.

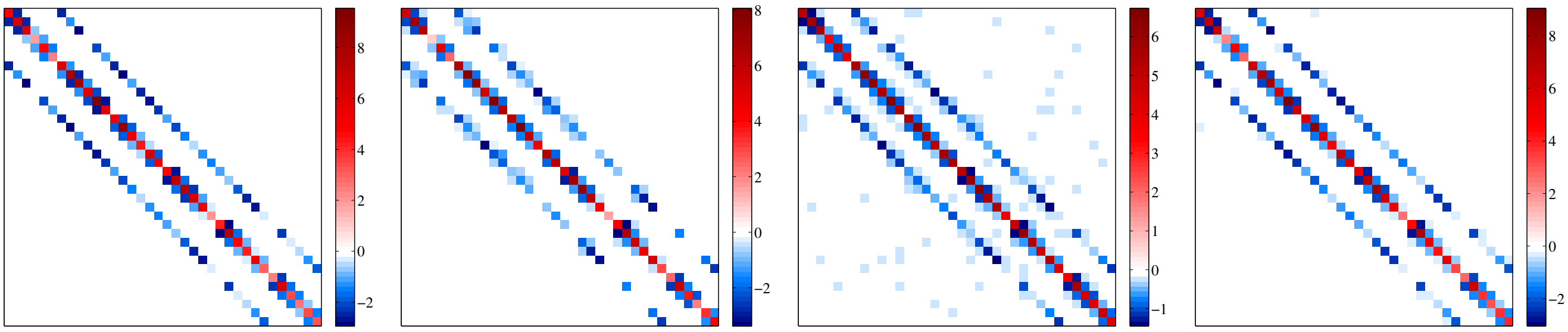

(a) The ground truth CGL ($\boldsymbol{\Theta}^*$) (b) GLS: (RE,FS) = (0.187,0.75) (c) GTI: (RE,FS) = (0.322,0.39) (d) CGL: (RE,FS) = (0.085,0.82)

Fig. 5. Illustration of precision matrices (whose entries are color coded) estimated from $\mathbf{S}$ for $k/n = 30$: In this example, (a) the ground truth is a CGL associated with a grid graph having $n = 36$ vertices. From $\mathbf{S}$, the matrices are estimated by (b) GLS($\alpha_1$), (c) GTI and (d) CGL($\alpha$). The proposed method provides the best estimation in terms of RE and FS, and the resulting matrix is visually the most similar to the ground truth $\boldsymbol{\Theta}^*$.

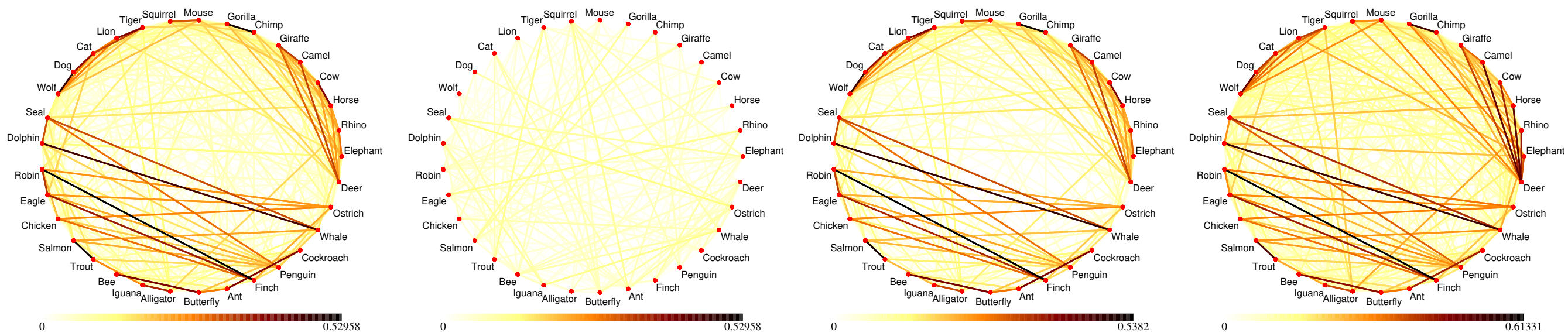

(a) GLasso($\alpha = 0$): (Left) Positive and (Right) absolute negative weights (b) GGL($\alpha = 0$) (c) CGL($\alpha = 0$)

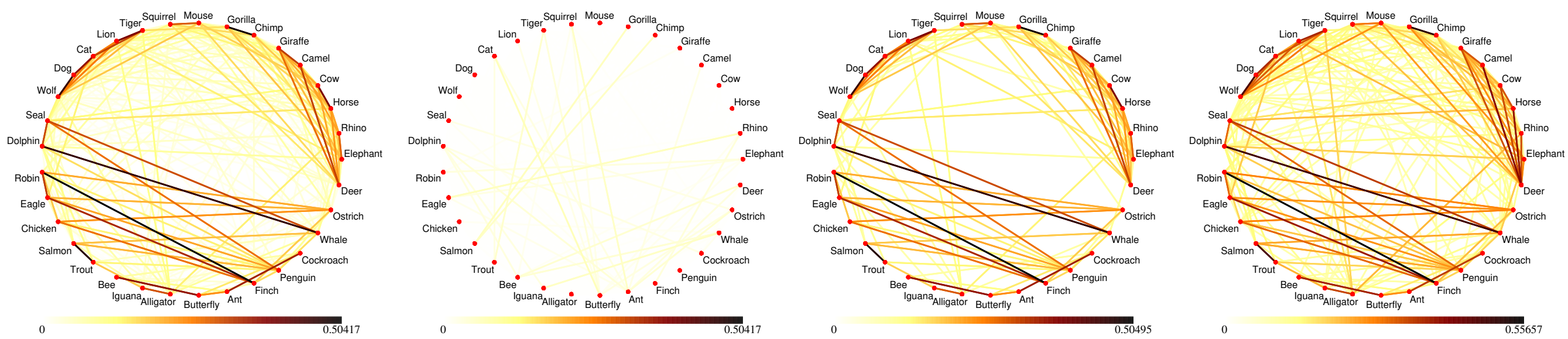

(d) GLasso($\alpha = 0.02$): (Left) Positive and (Right) absolute negative weights (e) GGL($\alpha = 0.02$) (f) CGL($\alpha = 0.02$)

Fig. 6. The graphs learned from Animals dataset using methods GLasso($\alpha$), GGL($\alpha$) and CGL($\alpha$): GGL($\alpha$) leads to sparser graphs compared to the others. The results follow the intuition that larger positive weights should be assigned between similar animals, although the dataset is categorical (non-Gaussian).

between them. The dataset consists of binary values (0 or 1) assigned to $d = 102$ features for $n = 33$ animals. Each feature corresponds to a true-false question such as, “has lungs?”, “is warm-blooded?” and “lives in groups?”. Using this dataset, the statistic matrix $\mathbf{S}$ is calculated as $\mathbf{S} = (1/d)\,\mathbf{X}\mathbf{X}^{\mathsf{T}} + (1/3)\,\mathbf{I}$ where $\mathbf{X}$ is the mean removed $n \times d$ data matrix. The $(1/3)\,\mathbf{I}$ term is added based on the variational Bayesian approximation result in [13] for binary data. Then, $\mathbf{S}$ is used as input to

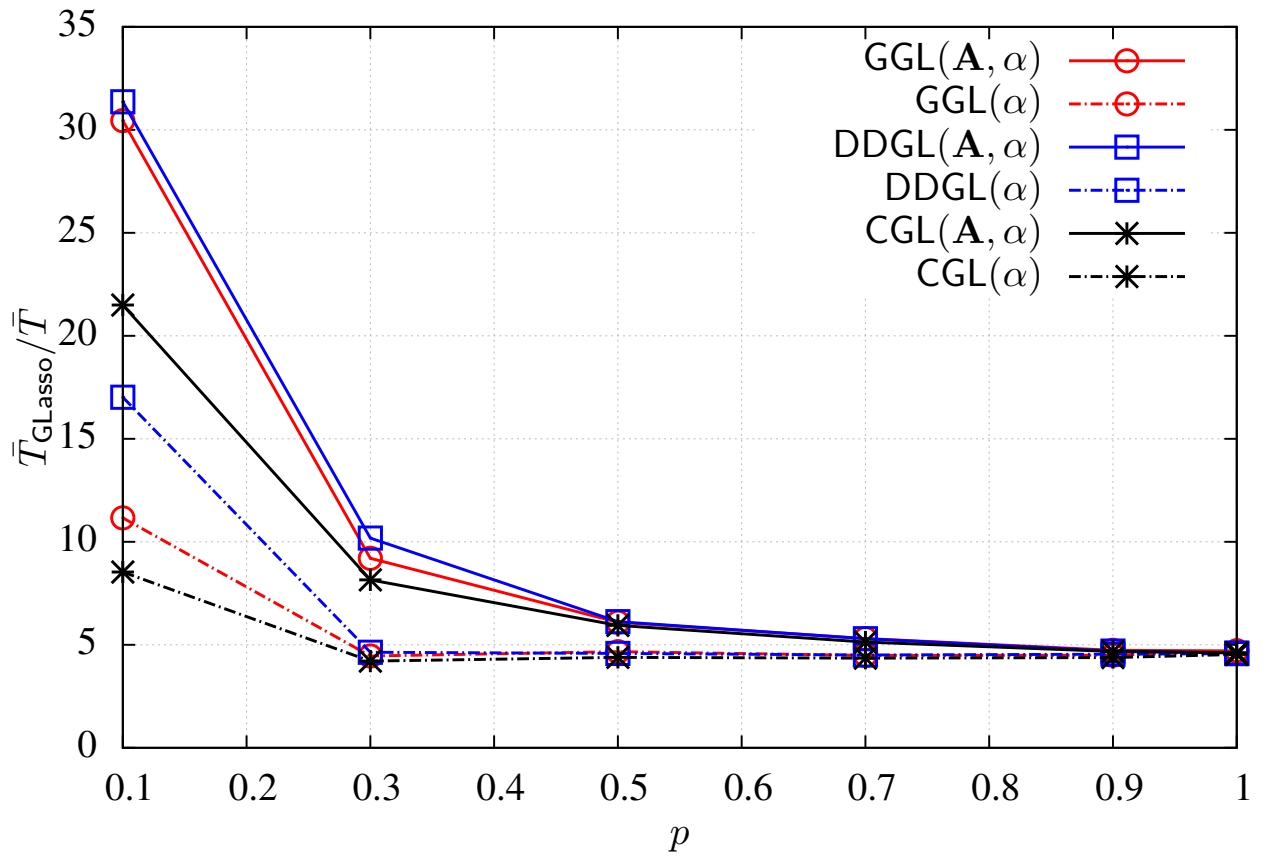

Fig. 7. Computational speedup as compared to GLasso ($\bar{T}_{\text{GLasso}}/\bar{T}$).

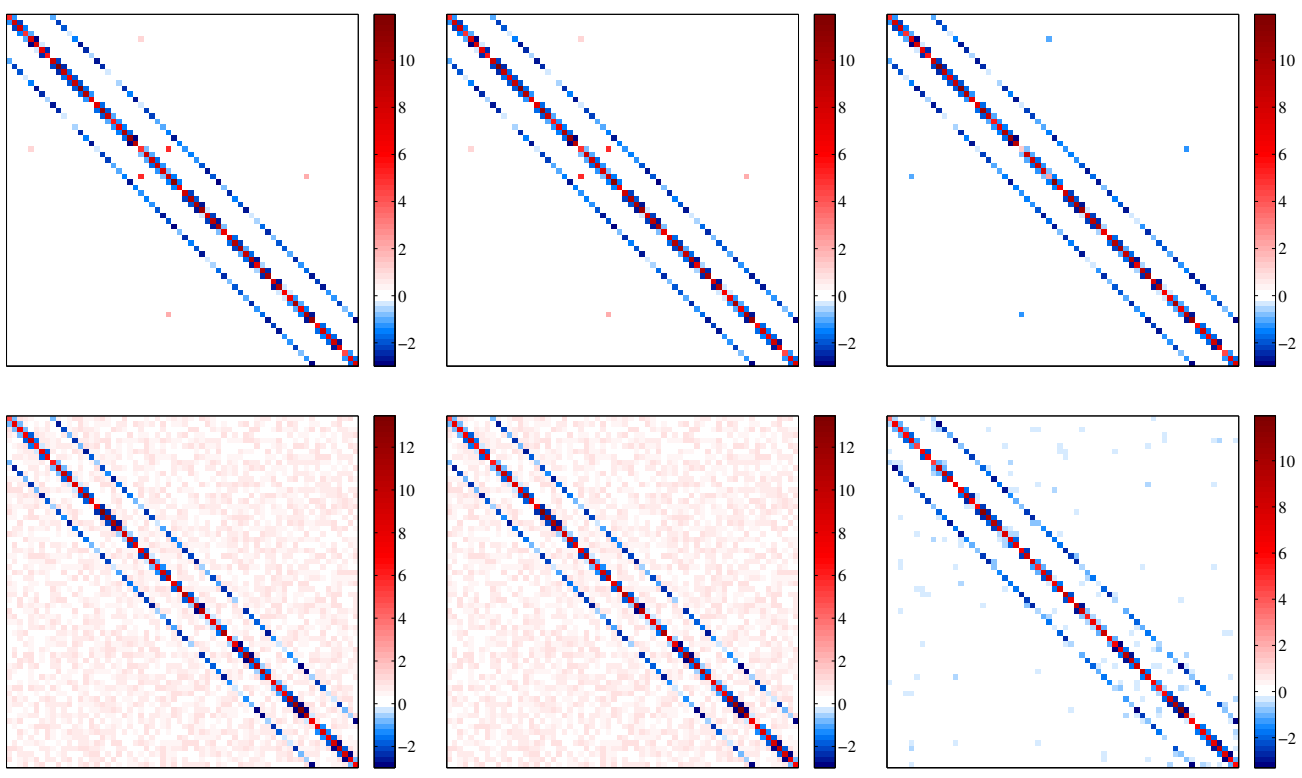

(a) Ground Truth ($\mathbf{\Omega}$) (b) GLasso($\alpha = 0$) (c) GGL($\alpha = 0$)

Fig. 8. Illustration of precision matrices whose entries are color coded, where negative (resp. positive) entries correspond to positive (resp. negative) edge weights of a graph: (a) Ground truth precision matrices ($\mathbf{\Omega}$) are randomly generated with (top) sparse and (bottom) dense positive entries (i.e., negative edge weights). From the corresponding true covariances ($\mathbf{\Sigma}$), the precision matrices are estimated by (b) GLasso($\alpha = 0$) and (c) GGL($\alpha = 0$) methods without $\ell_1$-regularization.

the GLasso, GGL and CGL methods. Here, we only compare methods minimizing the objective function in (1) which is shown to be a suitable metric for binary data in [13] [14].

Figure 6 illustrates the graphs constructed by using GLasso, GGL and CGL with different regularization parameters. The positive and negative edge weights estimated by GLasso are shown side-by-side in Figs. 6a and 6d, where the magnitudes of negative weights are substantially smaller than most of the positive weights. In other words, positive partial correlations are more dominant than negative partial correlations in the precision matrix. Since the proposed GGL and CGL find graphs with nonnegative edge weights (i.e., closest graph Laplacian projections with no negative edge weights), the corresponding results are similar to the graphs with nonnegative weights in Figs. 6a and 6d. The results follow the intuition that larger positive weights should be assigned between animals considered to be similar. Such pairs of animals include (gorilla,chimp), (dolphin,whale), (dog,wolf) and (robin,finch).

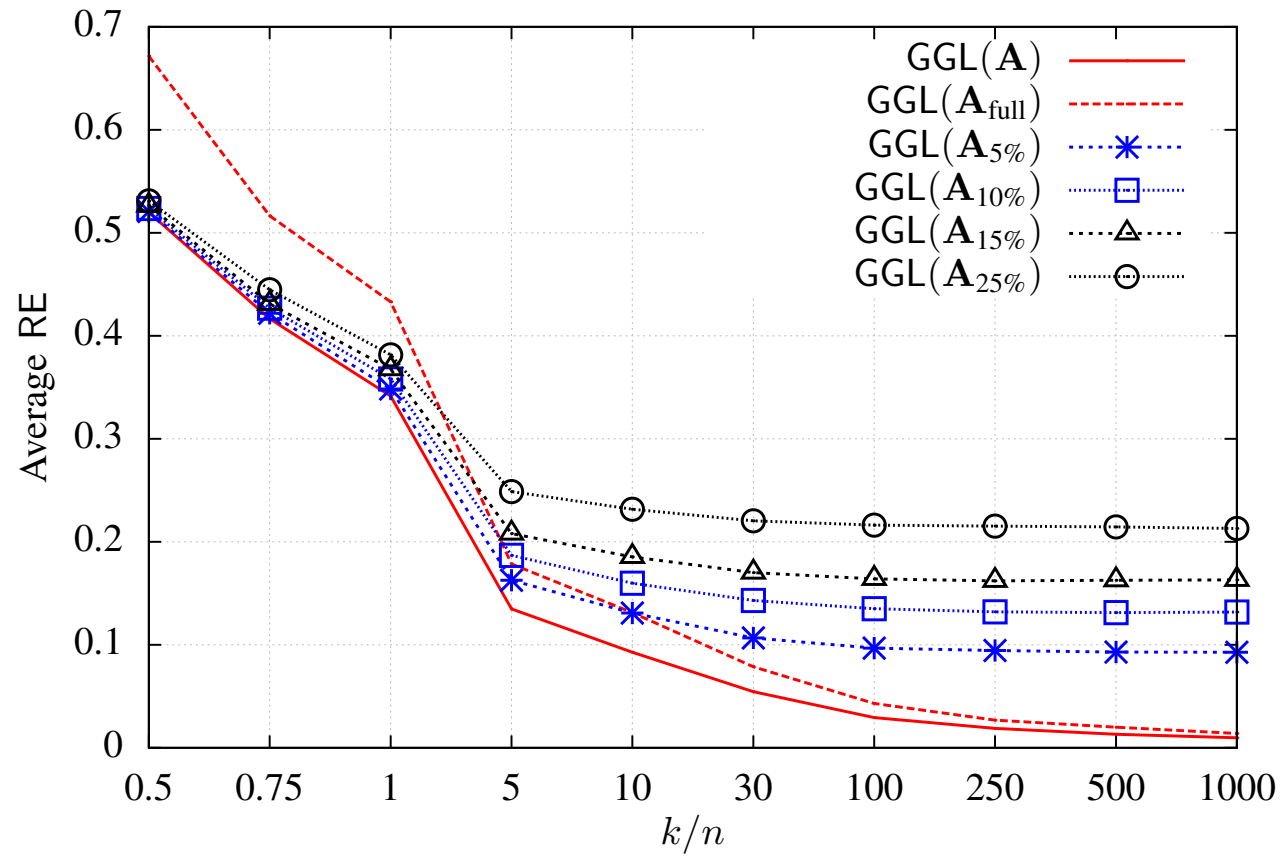

Fig. 9. Average relative errors for different number of data samples ($k/n$), where GGL($\mathbf{A}$) exploits the true connectivity information in GGL estimation, and GGL($\mathbf{A}_{\text{full}}$) refers to the GGL estimation without connectivity constraints. Different levels of connectivity mismatch are tested. For example, GGL($\mathbf{A}_{5\%}$) corresponds to the GGL estimation with connectivity constraints having a 5% mismatch. No $\ell_1$-regularization is applied in GGL estimation (i.e., $\alpha = 0$).

### *D. Graph Laplacian Estimation under Model Mismatch*

In the case of a model mismatch (i.e., when data is not generated by a GMRF with a graph Laplacian as its precision matrix), the proposed methods allow us to find the closest graph Laplacian fit with respect to the original model in the log-determinant Bregman divergence sense. Fig. 8 illustrates two examples (simulating sparse and dense mismatches) where the ground truth precision matrices $\mathbf{\Omega}$ have both negative and positive edge weights (Fig. 8a). From their true covariances $\mathbf{\Sigma} = \mathbf{\Omega}^{-1}$, GLasso recovers the original model by allowing both positive and negative edge weights (Fig. 8b). On the other hand, the proposed GGL finds the closest graph Laplacian with respect to the ground truth (Fig. 8c). As shown in the figure, GGL maintains all the edges with positive weights and also introduces additional connectivity due to the negative weights in the ground truth $\mathbf{\Omega}$. Note that this form of projection (based on log-determinant Bregman divergence) does not necessarily assign zeros to the corresponding negative edge weights. In general, the identification of edges with positive weights under model mismatches is nontrivial, as also pointed out in [39].

### *E. Graph Laplacian Estimation under Connectivity Mismatch*

We now evaluate empirically the effect inaccurate selection of connectivity matrices ($\mathbf{A}$), used to determine the connectivity constraints in our problems. For this purpose, in our experiments, we randomly construct multiple $64 \times 64$ GGL matrices based on Erdos-Renyi graphs with $p = 0.1$ (i.e., $\mathcal{G}_{\text{ER}}^{(64,0.1)}$) and generate datasets associated with the GGLs, as discussed in Section VI-A. Then, the proposed GGL method is employed to estimate graphs from generated data using different connectivity constraints, whose corresponding connectivity matrices are obtained by randomly swapping the entries of the true connectivity matrix to simulate connectivity mismatches. Specifically, a 5% mismatch is obtained by randomly exchanging 5% of the ones in $\mathbf{A}$ with zero entries.

Figure 9 compares the accuracy of GGL estimation with connectivity constraints under different levels (5%, 10%, 15% and 25%) of connectivity mismatch against the GGL estimation with true connectivity constraints (i.e., $\mathsf{GGL}(\mathbf{A})$) and without using any connectivity constraints (i.e., $\mathsf{GGL}(\mathbf{A}_{\text{full}})$). The results show that using slightly inaccurate connectivity information can be useful when the number of data samples is small (e.g., $k/n<5$), where the performance is similar to $\mathsf{GGL}(\mathbf{A})$ and can outperform $\mathsf{GGL}(\mathbf{A}_{\text{full}})$, even though there is a 25% mismatch ($\mathsf{GGL}(\mathbf{A}_{25\%})$). However, the performance difference with respect to $\mathsf{GGL}(\mathbf{A})$ increases as the number of data samples increases (e.g., $k/n>10$), where both $\mathsf{GGL}(\mathbf{A})$ and $\mathsf{GGL}(\mathbf{A}_{\text{full}})$ performs substantially better. In this case, the graph estimation without connectivity constraints ($\mathsf{GGL}(\mathbf{A}_{\text{full}})$) can be preferred if connectivity information is uncertain.

## VII. Conclusion

In this work, we have formulated various graph learning problems as estimation of graph Laplacian matrices, and proposed efficient block-coordinate descent algorithms. We have also discussed probabilistic interpretations of our formulations by showing that they are equivalent to parameter estimation of different classes of attractive GMRFs. The experimental results have demonstrated that our proposed techniques outperform the state-of-the-art methods in terms of accuracy and computational efficiency, and both can be significantly improved by exploiting the additional structural (connectivity) information about the problem at hand.

The methods proposed in this paper can be used in applications that involve learning a similarity graph, parameter estimation of graph-based models (e.g., attractive GMRFs) from data. Also, depending on the application, learning specific type of graph Laplacian matrices (generalized, diagonally dominant or combinatorial Laplacian matrices) can also be useful. Since the main focus of the present paper is on formulation of graph learning problems and development of new algorithms, related applications are considered as part of our future work.

## Appendix: Proofs for Propositions

*Proof of Proposition 1.* The problems in (9) and (10) have the same constraints. To prove their equivalence, we show that their objective functions are also the same. First, note that

$$\operatorname{Tr}(\boldsymbol{\Theta}(\mathbf{K}+\mathbf{J}))=\operatorname{Tr}(\boldsymbol{\Theta}\mathbf{K})+\frac{1}{n}\operatorname{Tr}(\boldsymbol{\Theta}\mathbf{1}\mathbf{1}^{\mathsf{T}})=\operatorname{Tr}(\boldsymbol{\Theta}\mathbf{K})$$

since $\boldsymbol{\Theta}\mathbf{1}=\mathbf{0}$ based on the CGL problem constraints. Next, we can write

$$\operatorname{logdet}(\boldsymbol{\Theta}+1/n\,\mathbf{1}\mathbf{1}^{\mathsf{T}})=\log\left(\prod_{i=1}^{n}\lambda_i(\boldsymbol{\Theta}+1/n\,\mathbf{1}\mathbf{1}^{\mathsf{T}})\right)\quad(63)$$

where $\lambda_i(\boldsymbol{\Theta})$ denotes the $i$-th eigenvalue of $\boldsymbol{\Theta}$ in ascending order ($\lambda_1(\boldsymbol{\Theta})\le\ldots\le\lambda_n(\boldsymbol{\Theta})$). Since the eigenvector corresponding to the first (zero) eigenvalue (i.e., $\lambda_1(\boldsymbol{\Theta})=0$) is $\mathbf{u}_1=1/\sqrt{n}\,\mathbf{1}$, by the problem constraints (i.e., by definition of CGL matrices), we have that

$$\boldsymbol{\Theta}+\frac{1}{n}\mathbf{1}\mathbf{1}^{\mathsf{T}}=(\underbrace{\lambda_1(\boldsymbol{\Theta})}_{0}+1)\mathbf{u}_1\mathbf{u}_1^{\mathsf{T}}+\sum_{i=2}^{n}\lambda_i(\boldsymbol{\Theta})\mathbf{u}_i\mathbf{u}_i^{\mathsf{T}}.\quad(64)$$

Since the determinant of a matrix is equal to the product of its eigenvalues, from (64) we have

$$\operatorname{logdet}(\boldsymbol{\Theta}+1/n\,\mathbf{1}\mathbf{1}^{\mathsf{T}})=\log\left(1\cdot\prod_{i=2}^{n}\lambda_i(\boldsymbol{\Theta})\right)=\log|\boldsymbol{\Theta}|.$$

Therefore, the problems in (9) and (10) are equivalent. □

*Proof of Proposition 2.* The function $\operatorname{logdet}(\boldsymbol{\Theta})$ defined over positive semidefinite matrices ($\boldsymbol{\Theta}\succeq 0$) is a concave function (see [56] for a proof), and $\operatorname{Tr}(\cdot)$ is a linear function. Thus, the overall objective function is convex. The graph Laplacian constraints form a convex set. Since we have a minimization of a convex objective function over a convex set, the problems of interest are convex. □

*Proof of Proposition 3.* The optimal $\nu$ cannot be zero, since the objective function is unbounded for $\nu=0$. By assuming that $\alpha=0$ (without loss of generality), the objective function in (12) can be decomposed as follows

$$\mathcal{J}(\widetilde{\boldsymbol{\Theta}})+\nu\operatorname{Tr}(\mathbf{S})-\sum_{i=2}^{n}\log\left(1+\frac{\nu}{\lambda_i(\widetilde{\boldsymbol{\Theta}})}\right)-\log(\nu)\quad(65)$$

where $\mathcal{J}(\widetilde{\boldsymbol{\Theta}})$ is the objective of Problem 3 with $\mathbf{H}=\mathbf{O}$. □

*Proof of Proposition 4.* For $\alpha_1=0$ and $\alpha_2=1$, the objective function in (14) is written as $\operatorname{Tr}(\boldsymbol{\Theta}\mathbf{S})-\sum_{i=1}^{n}\log((\boldsymbol{\Theta})_{ii})$. By using Hadamard's inequality $\det(\boldsymbol{\Theta})\le\prod_{i=1}^{n}(\boldsymbol{\Theta})_{ii}$ [64] and taking the $\log$ of both sides, the following bound is obtained

$$\operatorname{Tr}(\boldsymbol{\Theta}\mathbf{S})-\sum_{i=1}^{n}\log((\boldsymbol{\Theta})_{ii})\le\operatorname{Tr}(\boldsymbol{\Theta}\mathbf{S})-\operatorname{logdet}(\boldsymbol{\Theta})\quad(66)$$

where the right-hand side is the objective function in Problems 1, 2 and 3 with $\alpha=0$, as desired. □

*Proof of Proposition 5.* With proper choices of the prior $\mathsf{p}(\mathbf{L})$ in (23), the objective function in (1) can be constructed for any $\mathbf{H}$, except the pseudo-determinant term $|\boldsymbol{\Theta}|$ needed for estimating CGL matrices. For this case, Proposition 1 shows that we can equivalently formulate (23) in the form of Problem 3. The construction in (23)–(25) can be trivially extended for the weighted $\ell_1$-regularization. Also, the connectivity and Laplacian constraints in Problems 1, 2 and 3 can be incorporated in a Bayesian setting by choosing spike-and-slab prior and improper prior distributions [65] on $\mathbf{v}$ and $\mathbf{w}$, so that spike priors correspond to zero edge weights, and slab priors allow nonnegative edge weights. □

*Proof of Proposition 6.* In minimization of a strictly convex and differentiable function over a convex set, (with a proper selection of the step size) a block-coordinate descent algorithm guarantees convergence to the optimal solution [47], [55]. As stated in Proposition 2, all of our problems of interest are convex. Also, the objective functions are differentiable (see in Section V). Moreover, convergence conditions (see in [55]) require that each block-coordinate update be optimal (i.e., each iteration optimally solves the subproblem associated with selected coordinate directions). In Algorithms 1 and 2, this condition is also satisfied, since nonnegative quadratic programs (subproblems) are derived using optimality conditions, so each of their solutions leads to optimal block-coordinate (row/column) updates. These subproblems are also strictly convex, since the $\mathbf{Q}$ matrix in (44) and (56) is positive definite throughout all iterations. To prove this, we use the following Schur complement condition on partitions of $\mathbf{\Theta}$ as in (26),

$$\mathbf{\Theta} \succ 0 \iff \mathbf{\Theta}_u \succ 0 \text{ and } \theta_u - \boldsymbol{\theta}_u^{\mathsf{T}} \mathbf{\Theta}_u^{-1} \boldsymbol{\theta}_u > 0. \qquad (67)$$

where $\mathbf{\Theta}_u$ (and therefore, $\mathbf{\Theta}_u^{-1}$) is fixed and positive definite from the previous iteration. Noting that both algorithms initialize $\mathbf{\Theta}$ and $\mathbf{C}$ as (diagonal) positive definite matrices, $\theta_u - \boldsymbol{\theta}_u^{\mathsf{T}} \mathbf{\Theta}_u^{-1} \boldsymbol{\theta}_u = 1/c_u > 0$ by (29), then the updated $\mathbf{\Theta}$ is also positive definite. Since both algorithms solve for the optimal row/column updates at each iteration, based on the result in [55], this proves convergence to the optimal solution for Problem 1. For Problems 2 and 3, Algorithms 1 and 2 implement a variation of the general projected block-coordinate descent method, whose convergence to the optimal solution is shown in [58]. Also, it is trivial to show that additional updates on diagonal elements (see in (31)) maintain positive definiteness of $\mathbf{\Theta}$ and $\mathbf{C}$. Therefore, both algorithms guarantee convergence to the optimal solution. □